\pdfoutput=1

\documentclass[11pt]{article}

\usepackage[final]{acl}

\usepackage{times}
\usepackage{latexsym}
\usepackage{url}
\usepackage{amsmath}
\usepackage{graphicx}
\usepackage{booktabs}
\usepackage{multirow}
\usepackage{listings}
\usepackage{bbm}
\usepackage{subcaption}

\usepackage[T1]{fontenc}

\usepackage[utf8]{inputenc}

\usepackage{microtype}

\usepackage{inconsolata}

\usepackage{graphicx}
\usepackage[linesnumbered,ruled,vlined]{algorithm2e}

\usepackage{xspace,mfirstuc,tabulary}

\newcommand{\method}{\emph{MixSumm}}
\newcommand{\methodSSL}{\emph{PPSL}}

\newcommand{\lmodel}{\texttt{LLaMA-3-70b-Instruct}}
\newcommand{\dtrain}{$\mathcal{D}_{F, train}$}
\newcommand{\daug}{$\mathcal{D}_{A, train}$}
\newcommand{\dall}{$\mathcal{D}_{F+A, train}$}
\newcommand{\dulbl}{$\mathcal{D}_{U, train}$}
\newcommand{\stdfmt}[1]{{\scriptsize \textcolor{gray}{$\pm$#1}}}
\newcommand{\yellow}[1]{{\colorbox{yellow}{#1}}}
\newcommand{\cyan}[1]{{\colorbox{cyan}{#1}}}

\definecolor{codegreen}{rgb}{0,0.6,0}
\definecolor{codegray}{rgb}{0.5,0.5,0.5}
\definecolor{codepink}{RGB}{252, 142, 172}
\definecolor{codepurple}{rgb}{0.58,0,0.82}
\definecolor{backcolour}{RGB}{245,245,245}
\lstdefinestyle{mystyle}{
    backgroundcolor=\color{backcolour},   
    commentstyle=\color{magenta},
    keywordstyle=\color{blue},
    numberstyle=\tiny\color{codegray},
    stringstyle=\color{codepurple},
    basicstyle=\fontfamily{\ttdefault}\footnotesize,
    breakatwhitespace=false,         
    breaklines=true,                 
    keepspaces=true,    
    frame=single,
    numbersep=5pt,                  
    showspaces=false,                
    showstringspaces=false,
    showtabs=false,                  
    tabsize=2,
    classoffset=1, 
    keywordstyle=\color{violet},
    classoffset=0,
}
\lstdefinelanguage{JavaScript}{
  keywords={typeof, new, true, false, catch, function, return, null, catch, switch, var, if, in, while, do, else, case, break},
  keywordstyle=\color{blue}\bfseries,
  ndkeywords={class, export, boolean, throw, implements, import, this},
  ndkeywordstyle=\color{darkgray}\bfseries,
  identifierstyle=\color{black},
  sensitive=false,
  comment=[l]{//},
  morecomment=[s]{/*}{*/},
  commentstyle=\color{purple}\ttfamily,
  stringstyle=\color{red}\ttfamily,
  morestring=[b]',
  morestring=[b]"
}

\title{A Guide To Effectively Leveraging LLMs for Low-Resource Text
Summarization: Data Augmentation and Semi-supervised Approaches}

\author{
  Gaurav Sahu$^{1, 2}$ Olga Vechtomova$^{1}$ Issam H. Laradji$^{1,2,3}$
  \\
  $^{1}$Cheriton School of Computer Science,
  University of Waterloo, Canada \\
  $^{2}$ServiceNow Research \\
  $^{3}$University of British Columbia, Canada\\
  \texttt{gaurav.sahu@uwaterloo.ca}
}

\begin{document}
\maketitle
\begin{abstract}
Existing approaches for low-resource text summarization primarily employ large language models (LLMs) like GPT-3 or GPT-4 at inference time to generate summaries directly; however, such approaches often suffer from inconsistent LLM outputs and are difficult to adapt to domain-specific data in low-resource scenarios.
In this work, we propose two novel methods to effectively utilize LLMs for low-resource text summarization: \textbf{1)} {\method}, an LLM-based data augmentation regime that synthesizes high-quality documents (short and long) for \textit{few-shot text summarization}, and \textbf{2)} {\methodSSL}, a prompt-based pseudolabeling strategy for sample-efficient \textit{semi-supervised text summarization}.
Specifically, {\method} leverages the open-source {\lmodel} model to generate new documents by mixing topical information derived from a small seed set, and {\methodSSL} leverages the {\lmodel} model to generate high-quality pseudo-labels in a semi-supervised learning setup.
We evaluate our methods on the TweetSumm, WikiHow, and ArXiv/PubMed datasets and use L-Eval, a LLaMA-3-based evaluation metric, and ROUGE scores to measure the quality of generated summaries.
Our experiments on extractive and abstractive summarization show that {\method} and {\methodSSL} achieve competitive ROUGE scores as a fully supervised method with  5\% of the labeled data.
\end{abstract}

\section{Introduction}
\begin{figure}[t]
	\centering
	\includegraphics[width=\linewidth]{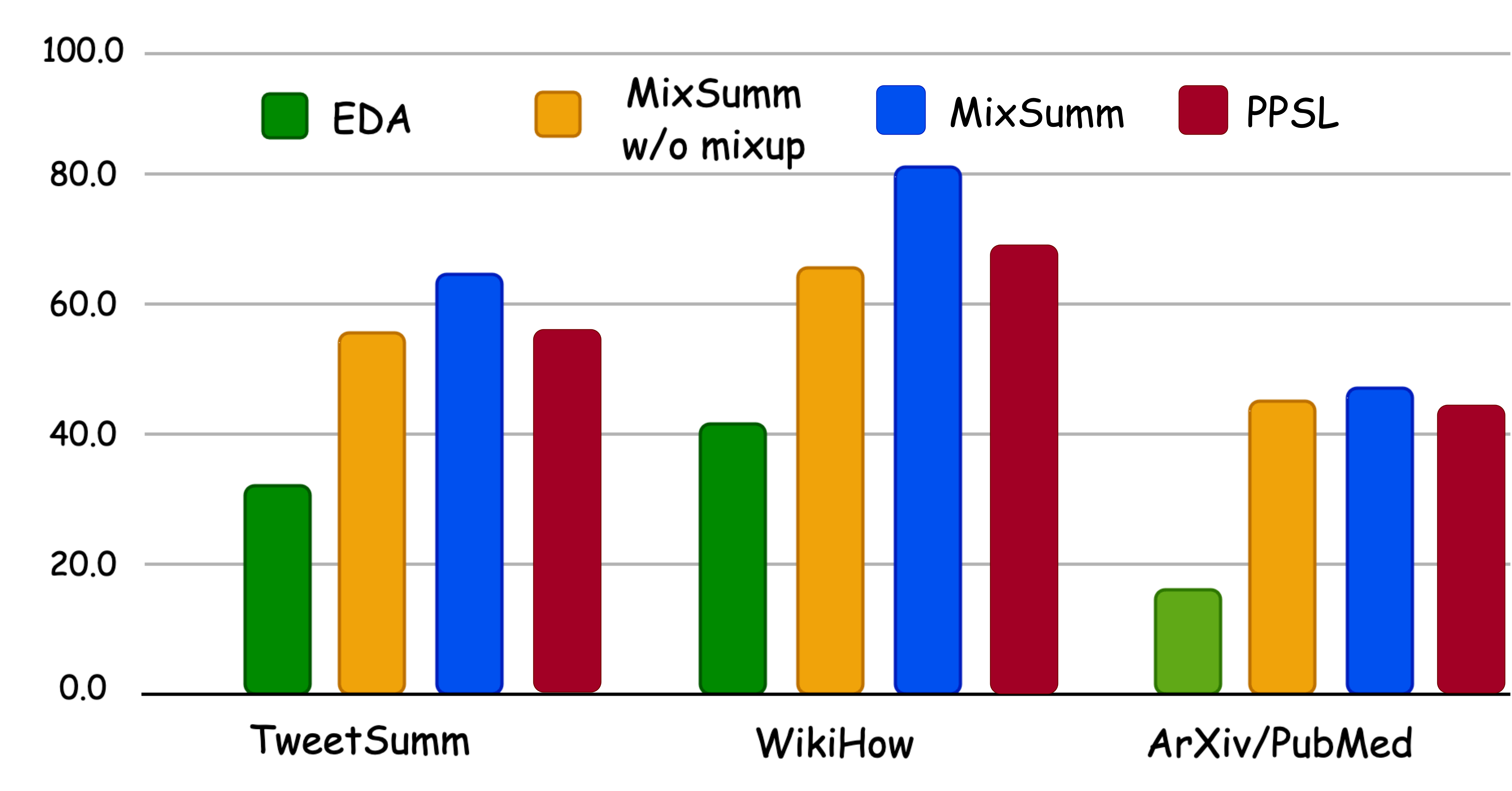}  
	\caption{\textbf{L-Eval scores of different methods on low-resource extractive text summarization.} The proposed {\method} approach generates new documents by combining topics from multiple examples and outperforms other methods, including a strong LLM-based DA method ({\method} w/o mixup) and a prompt-based semi-supervised approach (PPSL).
 }
	\label{fig:teaser}
\end{figure}

\begin{figure*}[t!]
    \centering
    \includegraphics[width=0.95\linewidth]{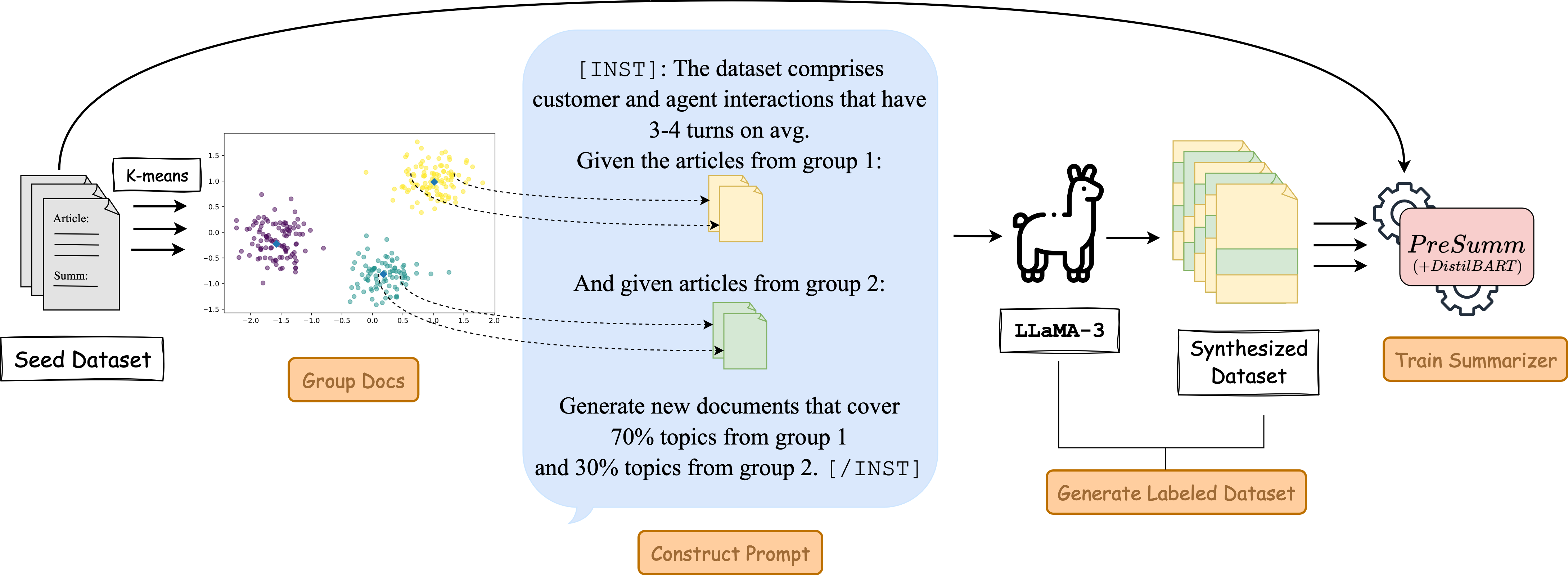}
    \caption{\textbf{{\method} pipeline.} We first group the documents into $T$ groups using the $k$-means algorithm. Then, we construct the prompt for {\lmodel} by including documents from different groups and instructing the LLM to mix information from multiple topics when generating the new documents. Finally, we train a PreSumm extractive summarizer~\cite{liu-lapata-2019-text} on the combined seed and the synthesized dataset. For abstractive summarization, we add a DistilBART model after PreSumm.}
    \label{fig:fig1}
\end{figure*}

Text summarization is a crucial task in today's data-driven era, with applications ranging from news digests to summarizing scientific papers to summarizing customer chatlogs in enterprises~\citep{cohan2017scientific,zhong2020extractive,goyal2022news,feigenblat-etal-2021-tweetsumm-dialog}.
Modern summarization systems can be broadly categorized into two types: abstractive, where the generated summaries are concise \textit{paraphrases} of the source text~\cite{abs1,abs2}, and extractive, which select and arrange existing sentences in the source text~\cite{wong2008extractive,kryscinski2019evaluating}.
While abstractive methods produce more fluent and natural-sounding summaries--particularly beneficial for longer documents, extractive methods are valued for their simplicity and reliability in preserving factual accuracy; however, the performance of these summarization systems is often constrained by the availability and diversity of training data.

Data augmentation (DA) has been successfully used to address data scarcity, mitigate data annotation costs, and enhance model robustness in various natural language processing (NLP) tasks like text classification, summarization, and grammatical error correction~\cite{wei-zou-2019-eda,feng2021survey,wang2022noise}. 
Traditional augmentation methods involving synonym replacement, sentence shuffling, and back-translation are effective to some extent, but they quickly saturate as they do not fully capture the semantic nuances of the text; however, the recent surge in the development of LLMs like GPT-4~\citep{achiam2023gpt}, LLaMA-3~\cite{touvron2023llama}, and Claude-3~\citep{anthropic2024claude}, has given birth to the paradigm of LLM-based data augmentation techniques~\citep{dai2023auggpt,ding2024data} that can generate contextually rich textual augmentations to enhance the performance of various NLP models such as dialog modeling systems~\cite{chintagunta-etal-2021-medically,wan-etal-2022-unified} and text classifiers~\cite{yoo2021gpt3mix,sahu-etal-2022-data}.
Real-life scenarios also often have a small labeled set alongside a large pool of unlabeled data, and semi-supervised learning (SSL) has been successfully used in such scenarios for images and text classification~\citep{img1,img2,img3,text1,text2}.

Despite the recent advancements in LLMs, neither LLM-based DA methods nor LLM-based semi-supervised methods have been extensively explored for low-resource text summarization.
Therefore, in this work, we propose two novel methods to effectively utilize LLMs for low-resource text summarization: \textbf{1)} {\method}, an LLM-based data augmentation technique for few-shot text summarization, and \textbf{2)} Prompt-based Pseudo-labeling for Semi-supervised Learning ({\methodSSL}), a pseudo-labeling strategy for sample-efficient semi-supervised text summarization.
More specifically, {\method} is a two-stage prompt-based data augmentation approach that
\textbf{first} instructs an LLM to synthesize diverse documents by mixing topical information derived from a small set of seed documents, and \textbf{then} generates summaries for the synthesized documents.
On the other hand, {\methodSSL} is a multi-step pseudo-labeling strategy for semi-supervised learning that generates high-quality pseudo-labels and selects most informative samples in an SSL pipeline.

To evaluate the effectiveness of our proposed framework, we conduct extensive experiments on the TweetSumm~\citep{feigenblat-etal-2021-tweetsumm-dialog}, the WikiHow~\citep{koupaee2018wikihow}, and the ArXiv/PubMed~\citep{cohan-etal-2018-discourse} text summarization datasets.
We use the open-source {\lmodel} LLM for our tasks instead of a closed-source LLM like the GPT family of LLMs.
For evaluation, we use the standard ROUGE scores~\citep{lin-2004-rouge} as well as L-Eval, an open-source version of the promising LLM-based evaluator for text summarization, G-Eval~\citep{liu-etal-2023-g}.
Our experiments demonstrate that {\method} and {\methodSSL} outperform strong data augmentation and semi-supervised baselines for low-resource summarization setups \textit{and} we show a knowledge distillation effect, where the knowledge of a LLaMA-3-70b model is distilled into the a much smaller summarization model using BERT$_{base}$ and DistilBART backend (with 110M and 306M parameters, respectively).

\begin{figure*}[t]
    \centering
    \includegraphics[width=0.95\linewidth]{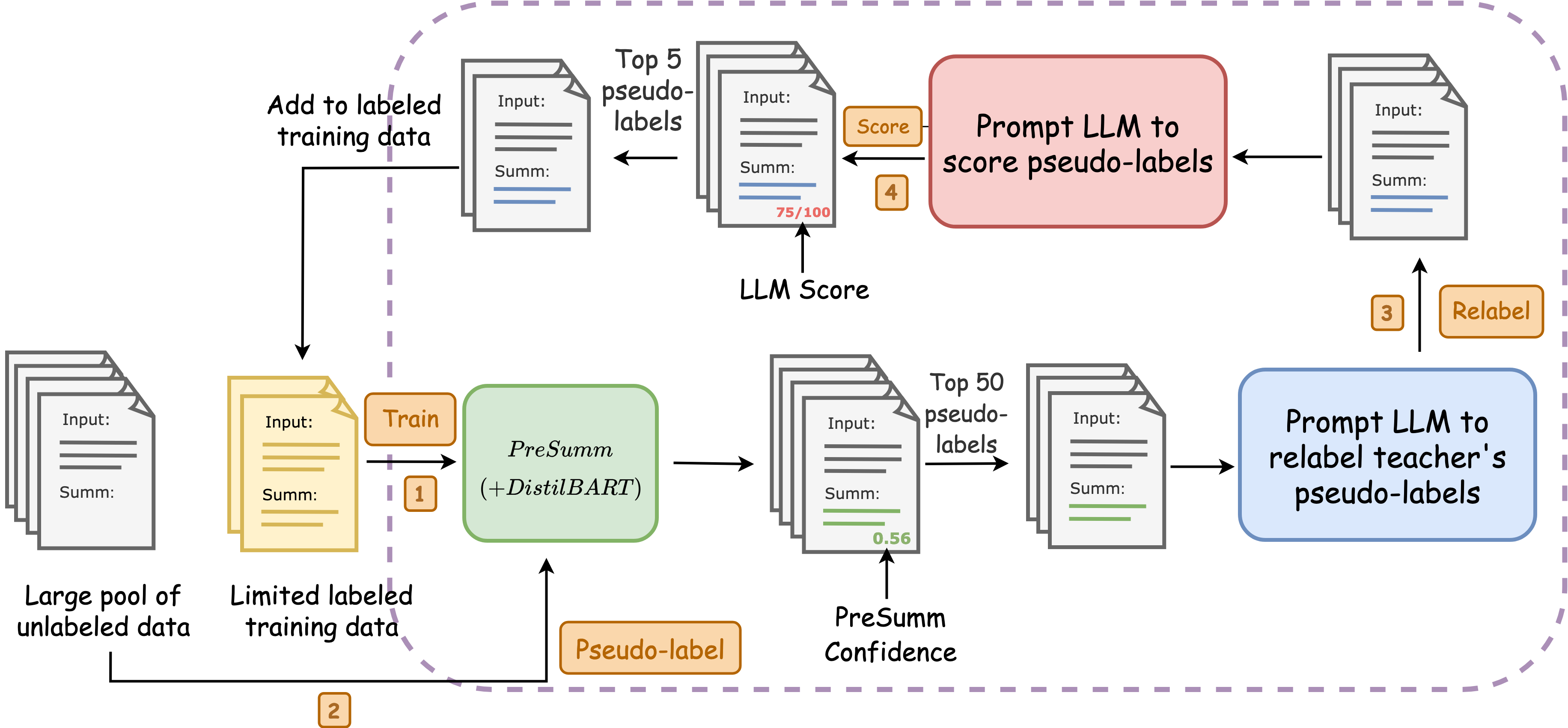}
    \caption{\textbf{{\methodSSL} pipeline.}  {\textbf{Step 1:}} train a teacher model $M$ on the limited labeled dataset. {\textbf{Step 2:}} generate pseudo-labels for the unlabeled set with $M$ and shortlist 50 based on teacher confidence (see Equation~\ref{eq:conf}). \textbf{Step 3:} prompt an LLM to summarize the shortlisted documents. \textbf{Step 4:} score the pseudo-labels in Stage 3 by prompting an LLM and select the top 5. These summaries are then added to the training data for the next cycle.}
    \label{fig:ssl_teaser}
\end{figure*}

To summarize the contributions of our work: \textbf{1)} we propose {\method}, a novel prompt-based data augmentation framework for the challenging low-resource setup of 50-shot text summarization, \textbf{2)} we propose {\methodSSL}, a novel pseudo-labeling strategy for sample-efficient semi-supervised text summarization, \textbf{3)} we show the effectiveness of {\lmodel}, an open-source LLM, instead of using expensive closed-source LLMs like GPT-4, and \textbf{4)} we demonstrate effective knowledge distillation from LLaMA-3-70B (70B parameters) to BERT and DistilBART-based summarization models with 110M and 306M parameters.
\section{Related Work}

\textbf{LLM-based Text Summarization.}
\citet{fabbri2020improving} use round-trip back-translation to improve BART's abstractive summarization performance.
On the other hand, \citet{dou2021gsum} propose GSum, a fully supervised transformer-based architecture that can use a guidance signal from an external source for improved abstractive text summarization.
\citet{goyal2022news} employ zero-shot prompting on GPT-3 for open-ended news summarization and show that humans overwhelmingly prefer GPT-3 summaries over human summaries.
\citet{pu2023chatgpt} use prompting on GPT-3 for controllable text summarization and show that while GPT-3 can follow simple constraints in the prompt like length, it shows a noticeably lower degree of change in styles compared to human-written summaries.
\citet{liu-etal-2024-benchmarking} and \citet{zhang2024benchmarking} benchmark the zero-shot performance of LLMs on instruction-controlled summarization and news summarization.
\citet{chintagunta-etal-2021-medically} use GPT-3 as a data annotator for 210-shot medical dialog summarization and show significant gains equivalent to using 6400 human-written labels.
More recently, \citet{liu-etal-2024-learning} fine-tune BART on LLM-generated summaries instead of human-generated summaries and show that LLMs are excellent references.
Notably, these works prompt GPT-3 directly for summarization in their experiments.
Except for the last two works, none of them use LLMs as data generators in low-resource setups.
Additionally, they all use a closed-source LLM in their experiments.
\citet{zhang2023extractive} propose an extract-then-generate method where they use in-context learning to generate extractive-summary-guided abstractive summaries.
However, since they operate in a fully-supervised setting, the method suffers from scalability issues for large datasets.
\citet{mishra2023llm} propose LLM pseudo-labeling for semi-supervised dialog summarization, but our proposed {\methodSSL} method is more sample-efficient as we use fewer labeled and unlabeled examples.

\noindent \textbf{LLM-based Distillation and Data Augmentation in NLP.}
A large body of recent work uses LLMs as data generators for distilling a large teacher model's knowledge into smaller models for training instruction-tuned models and chain-of-thought reasoning, while reducing human annotation load~\citep{ho-etal-2023-large,shum-etal-2023-automatic,meng2023tuning,liu-etal-2023-logicot,peng2023instruction}.
\citet{bonifacio2022inpars} use few-shot prompting to construct training datasets with query-document pairs for information retrieval.
In the landscape of few-shot text classification, \citet{yoo2021gpt3mix} propose GPT3Mix and \citet{sahu-etal-2022-data,sahu2023promptmix} propose PromptMix, where both the methods use LLMs as data generators and data labelers.
We are inspired by the success of LLM-based DA for these diverse NLP tasks and adopt the best prompting practices based on these works.
For instance, we generate diverse examples by mixing examples from different classes or groups as in GPT3Mix and PromptMix; and specify concrete criteria when using LLMs for generation and evaluation as in \citet{pu2023chatgpt} and \citet{liu-etal-2024-learning}.
Furthermore, we conduct extensive experiments to test the capabilities of an open-source LLM, LLaMA-3-70b, for low-resource text summarization, instead of using closed-source LLM like GPT-3 and GPT-4.
Finally, we also test our LLM-based DA strategy on extremely long documents.
\section{Notations}
\label{sec:notations}

We denote an annotated, many-shot summarization dataset as $\mathcal{D} = (d_i, s_i) \ \forall \ i \in \{1,\dots,N\}$, where $(d_i, s_i)$ denotes the $i$-th datapoint with input text document $d_i$ and its ground truth summary $s_i$.
We refer to the training, validation, and testing parts of the dataset as $\mathcal{D}_{train}$, $\mathcal{D}_{val}$, and $\mathcal{D}_{test}$, respectively.
Given the many-shot training set $\mathcal{D}_{train}$, we construct a few-shot version of the dataset with $k$ examples {\dtrain} and the unlabeled set {\dulbl} as follows:

\paragraph{Step 1.} Given {\dtrain}, we group the training articles by topics.
We do not define the topics explicitly and identify $T$ groups by applying the $k$-means algorithm on the document embeddings (where $k=T$)\footnote{to clarify, $k$ in $k$-means is different from $k$ in $k$-shot}.
We use the SBERT encodings~\citep{reimers-gurevych-2019-sentence} of the input documents as document embeddings\footnote{\texttt{sentence-transformers/all-mpnet-base-v2} model was used from the sentence-transformers library}.
If an input document exceeds SBERT's context window length of 512 tokens (roughly 300-400 English words), we chunk the document into smaller pieces and then average the chunk embeddings to obtain the final document embedding.

\paragraph{Step 2.} We construct our $k$-shot dataset {\dtrain} by randomly sampling an equal number of datapoints from each of the $T$ clusters so that {\dtrain} has $k$ examples in total.
In Section~\ref{sec:results}, we empirically show that our principled approach for constructing few-shot datasets is better than randomly sampling $k$ examples from $\mathcal{D}_{train}$ as it provides better topical coverage.
\paragraph{Step 3.} We randomly select $m$ documents from $\mathcal{D}_{train} \setminus \mathcal{D}_{F, train}$ (without labels) to construct the unlabeled set of documents {\dulbl} to be used in the semi-supervised setup.

\paragraph{Problem Formulation (few-shot setup).} Given a text summarization dataset $\mathcal{D}$: \textbf{1)} perform data augmentation on {\dtrain} to synthesize a labeled dataset {\daug}, and \textbf{2)} train a text summarization model on the combined dataset {\dall}.

\paragraph{Problem Formulation (semi-supervised setup).} Given a text summarization dataset $\mathcal{D}$: \textbf{1)} perform SSL on {\dtrain} and {\dulbl} to obtain a pseudo-labeled dataset $\mathcal{D}_{F+U, train}$, and \textbf{2)} train a text summarization model on the combined dataset $\mathcal{D}_{F+U, train}$.

\section{Methodology}
\label{sec:method}
\subsection{{\method} for Few-Shot Text Summarization}
We now describe, {\method}, a two-step approach for synthesizing labeled summarization documents.
\textbf{First}, instruct an LLM to generate documents that cover multiple topics derived from a small set set.
\textbf{Next}, we instruct an LLM to generate summaries for those documents.
The following sections describe our two-step procedure in detail.

\paragraph{Step 1: Synthesizing New Documents.}
\label{ssec:generate_docs}
\textit{First,} for every dataset, we manually write a short description that describes the type and approximate size of articles in the dataset.
These descriptions enable our approach to be used in even zero-shot settings.
\textit{Next,} we construct $T$ pairs of clusters $(c_i, c_j) \ \forall \ i, j \in {1,\dots,T}, i \ne j$, such that $c_j$ is the most distant cluster from $c_i$.
We use the centroids of the clusters obtained during $k$-means clustering in Section~\ref{sec:notations} for our computation.
We also ensure that all cluster pairs are \textit{unique} as $(c_i, c_j) \equiv (c_j, c_i)$.
\textit{Finally,} we combine the dataset description with $k$ examples from each cluster and instruct the LLM to generate new examples that cover topics from both clusters.
Specifically, we instruct the LLM to generate examples that contain $\alpha\%$ topics from the first cluster $c_i$ and $(1-\alpha)\%$ topics from the second cluster $c_j$, where $\alpha$ is sampled from a uniform distribution between 1-100.
This is similar to applying the mixup algorithm~\citep{zhang2018mixup} in a natural language space and has proven highly effective for data augmentation in low-resource text classification setups~\citep{yoo2021gpt3mix,sahu2023promptmix}.
Prompt~\ref{prompt:generate} in Appendix~\ref{prompt-design} shows the complete template for this step.

\paragraph{Step 2: Generating Summaries for the Synthesized Documents.}
Next, we instruct the LLM to generate extractive and abstractive summaries for the synthesized documents.
\textbf{For extractive summaries,} we provide a generated document to the LLM and then instruct it to output a probability score for each sentence indicating whether that sentence should be included in the summary or not.
We then rank the lines by the scores and choose the top-$p$ lines, where $p$ is the summary size and depends on the dataset.
We truncate the input document if it exceeds the LLM's context window length.
This approach ensures the extractiveness of the generated summary labels as it mimics PreSumm~\citep{liu-lapata-2019-text}, a strong baseline for extractive text summarization.
\textbf{For abstractive summaries,} instead of passing the entire source document and prompting the LLM to generate a summary, we ask it to summarize the previously generated extractive summaries.
This approach is faster than passing the source document and summarizing as our input context is significantly smaller.
More importantly, it enhances the factual correctness of the summaries.

\subsection{{\methodSSL} for Semi-Supervised Text Summariation}
This section describes our approach for semi-supervised text summarization.
As shown in Figure~\ref{fig:fig1}, we employ a teacher-student training framework and divide our pipeline into four steps, where we first train a teacher model on {\dtrain}, then use it to generate pseudo-labels for {\dulbl}, prompt the LLM to relabel the teacher's pseudo-labels from the previous step, and lastly score the new pseudo-labels with an LLM and select top 5 to include in the next training cycle.

\paragraph{Step 1: Training the Teacher Model}
First, we train a fully-supervised model $M$ (teacher) on the set of available labeled examples {\dtrain}.
We use PreSumm~\citep{liu-lapata-2019-text} as our extractive summarizer as it has been shown to perform well for extractive summarization.
Notably, PreSumm reformulates the task of generating extractive summaries to binary classification, where, for each sentence in the input document, the model predicts if it will be present in the output summary.
Then, the model combines the top-$n$ sentences with the highest probabilities in their order of appearance in the original text to construct the extractive summary.
\textbf{For abstractive summarization}, we follow an \textit{extractive-then-abstractive} approach and add a DistilBART model that summarizes PreSumm's summary. The rest of the subsequent steps remain unchanged.

\paragraph{Step 2: Generating Pseudo-labels using the Teacher Model}
We use the teacher model $M$ to generate pseudo-labels for the unlabeled set {\dulbl}.
Next, we shortlist a subset of 50 pseudo-labels with the highest teacher confidence~\footnote{We experiment 5, 10, 20, 50, and 75 and find 50 to be optimal.}.
We describe confidence computation in detail in Appendix~\ref{sec:ppsl_conf}
We will show in Section~\ref{sec:results} that shortlisting a subset of pseudo-labels helps make our method more sample-efficient, as we avoid relabeling a large unlabeled pool. This ultimately minimizes our LLM usage cost in the subsequent steps.

\paragraph{Step 3: LLM Relabeling of Teacher's Pseudo-labels}
After selecting the top 50 pseudo-labels using teacher confidence defined in Equation~\ref{eq:conf}, we prompt the LLM to generate a summary for each shortlisted unlabeled example.
This effectively relabels the pseudo-label from Step 2.
Specifically, we follow the prompt template in Figure~\ref{fig:gptrelabel} when generating summaries, which uses the same mechanism as the teacher $M$, i.e., for extractive summarization, we instruct the LLM to output probabilities for each sentence in the input document and then concatenate the top-$n$ lines in their order of appearance in the input text and for abstractive, we further ask the LLM to summarize the extractive summary.

\paragraph{Step 4: LLM Scoring of Pseudo-labels}
\label{sec:score}
In the last step of our {\methodSSL}, we prompt LLaMA-3 as shown in Figure~\ref{fig:gpt4score} to output a rating between 0-100 for the pseudo-labels from Step 3.
Finally, we choose the top 5 pseudo-labeled examples with the highest LLM scores and add them to the existing labeled set.
We repeat Steps 1-4 $N_{cycles}$ times to improve the initial summarization model $M$ and use the model obtained after the last cycle for generating summaries for the unseen test set.


\newcommand{\highlight}[1]{\colorbox{gray!30}{#1}}
\section{Experimental Setup}
\label{sec:exp_setup}
\subsection{Datasets}
We use three popular datasets in this work for extractive text summarization.

\noindent \textbf{1)} \textbf{TweetSumm}~\citep{feigenblat-etal-2021-tweetsumm-dialog} is a real-world customer service dataset that has 1100 conversations between a customer and an agent, and each conversation has three human-annotated extractive summaries.
The training set has 858 dialogs, and the validation and test sets have 100 examples each.
\textbf{2) WikiHow} \citep{koupaee2018wikihow} contains WikiHow articles with their headlines as abstractive summaries. The dataset has over 180k articles, with around 168k training articles and 6000 test and validation articles.
\textbf{3) ArXiv/PubMed} \citep{cohan-etal-2018-discourse} is a collection of scientific articles from PubMed and ArXiv with their abstracts as summaries.
The dataset has $\sim$325k articles, with nearly 300k training articles and 12.5k test and validation articles.

\begin{table}[t]
    \centering
    \resizebox{\linewidth}{!}
    {
    \begin{tabular}{lcccc}
         \toprule
         & TweetSumm & WikiHow & ArXiv/PubMed \\
         \midrule
         \# Train & 858 & 168,000 & 300,000 \\
         \# Valid & 100 & 6,000 & 12,500 \\
         \# Test & 100 & 6,000 & 12,500 \\
         Avg. Doc. Length & 245.01 & 579.8 & 4203.4 \\
        \bottomrule          
    \end{tabular}
    }
    \caption{Statistics of the text summarization datasets used in our experiments. \textbf{Note:} Avg. doc. length is reported in the number of tokens.}
    \label{tab:datastats}
\end{table}

Table~\ref{tab:datastats} summarizes the dataset statistics.
Since WikiHow and ArXiv/PubMed datasets do not have extractive labels, we follow the same steps as the original PreSumm paper~\citep{liu-lapata-2019-text} and construct an extractive summary that maximizes the ROUGE score between the obtained extractive summary and the ground-truth abstractive summary.
We chose the three datasets above as they cover diverse scenarios, from relatively short real-world customer-agent conversations in the TweetSumm dataset to long scientific articles in the ArXiv/PubMed dataset.
We report the training implementation details in Appendix~\ref{ssec:impl_details}.

\subsection{Evaluation.} We evaluate the summary quality of the models using the following metrics:

\paragraph{ROUGE Scores.} We use ROUGE-1 (R-1), ROUGE-2 (R-2), and ROUGE-L (R-L) F1 scores~\cite{lin-2004-rouge} for evaluation, where R-1 and R-2 measure the unigram and bigram overlap between the predicted and the ground truth summaries, respectively, while R-L also considers the order of $n$-grams.
We use the \texttt{pyrouge} Python package to compute ROUGE scores in our setup and report them in Table~\ref{tab:full_results}.

\begin{table*}[t]
    \centering
    \resizebox{0.9\linewidth}{!}
    {%
    \begin{tabular}{lcccccccccccc}
        \toprule
         & \multicolumn{4}{c}{\textbf{TweetSumm}} & \multicolumn{4}{c}{\textbf{WikiHow}} &
         \multicolumn{4}{c}{\textbf{ArXiv/Pubmed}} \\
         \cmidrule(r){2-13}
         \textbf{Method} & \textbf{R-1 (\%)} & \textbf{R-2 (\%)} & \textbf{R-L (\%)} & \textbf{L-Eval (\%)} & \textbf{R-1 (\%)} & \textbf{R-2 (\%)} & \textbf{R-L (\%)} & \textbf{L-Eval (\%)} & \textbf{R-1 (\%)} & \textbf{R-2 (\%)} & \textbf{R-L (\%)} & \textbf{L-Eval (\%)} \\
        \toprule
        \multicolumn{13}{c}{\textit{Extractive Summarization}}\\
        \midrule
        \textit{Oracle} & \textit{65.7\stdfmt{0.3}} & \textit{56.6\stdfmt{0.4}} & \textit{64.9\stdfmt{0.4}} & \textit{86.2\stdfmt{0.3}} & \textit{30.5\stdfmt{0.5}} & \textit{8.7\stdfmt{0.3}} & \textit{19.2\stdfmt{0.6}} & \textit{87.3\stdfmt{0.1}} & \textit{34.6\stdfmt{0.4}} & \textit{12.4\stdfmt{0.2}} & \textit{19.6\stdfmt{0.4}} & \textit{78.1\stdfmt{0.5}} \\
        \toprule
        TSL (50:500) & 49.0 & 37.7 & 48.2 & - & - & - & - & -  & - & - & - & - \\
        TSL (500:500) & 59.0 & 48.3 & 58.2 & - & - & - & - & - & - & - & - & - \\
        EDA & {51.1\stdfmt{0.7}} & {39.2\stdfmt{0.9}} & {53.0\stdfmt{0.2}} & 34.3\stdfmt{1.2} & {23.4\stdfmt{0.5}} & {4.1\stdfmt{0.3}} & {13.0\stdfmt{0.5}} & 42.1\stdfmt{0.8} & {26.2\stdfmt{1.1}} & {7.9\stdfmt{1.0}} & {13.1\stdfmt{0.6}} & 17.2\stdfmt{0.5} \\
        PPSL (50:250) & {58.4\stdfmt{1.2}} & {50.1\stdfmt{0.3}} & {59.1\stdfmt{1.2}} & 56.3\stdfmt{0.9} & {26.0\stdfmt{0.2}} & {6.9\stdfmt{0.3}} & {15.1\stdfmt{0.2}} & 69.3\stdfmt{2.1} & {29.0\stdfmt{0.5}} & {9.4\stdfmt{0.7}} & {17.4\stdfmt{0.3}} & 49.3\stdfmt{1.4} \\
        {\method} (rand.) & {58.6\stdfmt{3.2}} & {50.6\stdfmt{2.1}} & {59.7\stdfmt{2.3}} & 60.3\stdfmt{0.9} & {26.4\stdfmt{1.0}} & {7.5\stdfmt{1.2}} & {15.8\stdfmt{0.2}} & 72.5\stdfmt{1.2} & {30.7\stdfmt{1.7}} & \highlight{\textbf{10.6\stdfmt{1.5}}} & \highlight{\textbf{18.5\stdfmt{1.1}}} & 48.4\stdfmt{1.1} \\
        \quad w/o Aug. & {49.4\stdfmt{0.7}} & {36.9\stdfmt{1.0}} & {49.0\stdfmt{0.2}} & 31.5\stdfmt{0.5} & {21.3\stdfmt{0.4}} & {3.2\stdfmt{0.4}} & {11.4\stdfmt{0.5}} & 34.2\stdfmt{1.5} & {23.4\stdfmt{1.1}} & {7.5\stdfmt{1.4}} & {12.3\stdfmt{0.8}} & 13.5\stdfmt{1.2} \\
        {\method} (ours) & \textbf{59.1\stdfmt{1.7}} & \textbf{52.7\stdfmt{1.6}} & \highlight{\textbf{60.5\stdfmt{1.3}}} & \textbf{65.3\stdfmt{1.2}} & \textbf{27.3\stdfmt{2.1}} & \highlight{\textbf{7.8\stdfmt{1.3}}} & \textbf{16.6\stdfmt{1.8}} & \highlight{\textbf{81.1\stdfmt{1.7}}} & {\textbf{31.2\stdfmt{1.2}}} & \highlight{\textbf{10.7\stdfmt{1.1}}} & \highlight{\textbf{18.3\stdfmt{1.1}}} & \highlight{\textbf{53.1\stdfmt{0.5}}} \\
        \quad w/o Mixup & {56.1\stdfmt{1.1}} & {47.3\stdfmt{1.2}} & {55.3\stdfmt{1.1}} & 57.3\stdfmt{0.5} & {25.7\stdfmt{1.4}} & {6.2\stdfmt{1.2}} & {14.7\stdfmt{0.7}} & 67.3\stdfmt{2.1} & {28.4\stdfmt{1.9}} & {8.3\stdfmt{1.3}} & {16.8\stdfmt{1.6}} & 52.3\stdfmt{1.2} \\
        \quad w/o Aug. & {50.1\stdfmt{0.6}} & {38.1\stdfmt{1.0}} & {49.9\stdfmt{0.6}} & 32.3\stdfmt{3.1} & {21.9\stdfmt{0.3}} & {3.5\stdfmt{0.2}} & {12.1\stdfmt{0.9}} & 33.3\stdfmt{1.7} & {24.1\stdfmt{0.9}} & {7.9\stdfmt{1.0}} & {12.7\stdfmt{0.5}} & 19.0\stdfmt{2.5} \\
        \midrule
        LLaMA-3 (0-shot) & {50.3\stdfmt{0.5}} & {47.7\stdfmt{0.4}} & {49.9\stdfmt{0.3}} & 52.3\stdfmt{1.2} & {12.2\stdfmt{0.2}} & {2.7\stdfmt{0.5}} & {8.1\stdfmt{0.4}} & 32.3\stdfmt{0.3} & {23.6\stdfmt{0.2}} & {4.6\stdfmt{0.7}} & {15.4\stdfmt{0.3}} & \textbf{38.4\stdfmt{0.5}} \\
        LLaMA-3 (1-shot) & {51.7\stdfmt{0.2}} & {49.2\stdfmt{0.3}} & {51.9\stdfmt{0.3}} & 58.7\stdfmt{1.1} & {14.3\stdfmt{0.2}} & {4.1\stdfmt{0.5}} & {10.6\stdfmt{0.2}} & 39.4\stdfmt{0.5} & \highlight{\textbf{32.6\stdfmt{0.4}}} & \textbf{6.5\stdfmt{0.7}} & \textbf{17.2\stdfmt{0.3}} & 38.3\stdfmt{1.8} \\
        LLaMA-3 (5-shot) & \highlight{\textbf{62.4\stdfmt{0.5}}} & \highlight{\textbf{54.3\stdfmt{0.7}}} & \highlight{\textbf{60.3\stdfmt{1.1}}} & \highlight{\textbf{67.5\stdfmt{0.6}}} & \highlight{\textbf{28.7\stdfmt{0.3}}} & {\textbf{7.5\stdfmt{0.9}}} & \highlight{\textbf{17.1\stdfmt{0.3}}} & {\textbf{71.3\stdfmt{0.4}}} & - & - & - & - \\
        \toprule
        \multicolumn{13}{c}{\textit{Abstractive Summarization}}\\
        \toprule
        \textit{Oracle} & \textit{44.7\stdfmt{0.2}} & \textit{20.1\stdfmt{0.4}} & \textit{36.8\stdfmt{0.2}} & \textit{72.3\stdfmt{0.6}} & \textit{28.7\stdfmt{0.3}} & \textit{6.2\stdfmt{0.7}} & \textit{13.6\stdfmt{0.4}} & \textit{78.4\stdfmt{0.8}} & \textit{28.4\stdfmt{0.2}} & \textit{10.2\stdfmt{0.4}} & \textit{15.8\stdfmt{0.8}} & \textit{64.3\stdfmt{0.5}} \\
        \toprule
        EDA & {41.5\stdfmt{1.2}} & {15.0\stdfmt{0.8}} & {32.2\stdfmt{1.1}} & 44.2\stdfmt{1.6} & 14.7\stdfmt{1.8} & 3.2\stdfmt{1.0} & 6.8\stdfmt{1.5} & 40.5\stdfmt{1.4} & 16.3\stdfmt{1.5} & 5.9\stdfmt{0.8} & 8.1\stdfmt{1.7} & 36.8\stdfmt{1.3} \\
        PPSL (50:250) & {42.7\stdfmt{1.5}} & {18.1\stdfmt{1.1}} & {33.8\stdfmt{1.3}} & 58.1\stdfmt{1.3} & \highlight{\textbf{26.9\stdfmt{1.8}}} & \highlight{\textbf{5.7\stdfmt{1.0}}} & \highlight{\textbf{12.1\stdfmt{1.5}}} & 62.1\stdfmt{1.4} & 26.7\stdfmt{1.5} & 9.5\stdfmt{0.8} & \highlight{\textbf{13.8\stdfmt{1.7}}} & 61.3\stdfmt{1.3} \\
        {\method} (ours) & \textbf{43.1\stdfmt{1.1}} & \textbf{18.4\stdfmt{1.5}} & {\textbf{34.7\stdfmt{1.0}}} & \textbf{62.3\stdfmt{1.4}} & {26.7\stdfmt{1.7}} & {{5.3\stdfmt{0.9}}} & {11.3\stdfmt{1.4}} & {\textbf{67.5\stdfmt{1.3}}} & \highlight{\textbf{27.1\stdfmt{1.4}}} & \highlight{\textbf{9.8\stdfmt{0.7}}} & 13.5\stdfmt{1.6} & \highlight{\textbf{61.4\stdfmt{1.2}}} \\
        \quad w/o Mixup & 37.5\stdfmt{1.0} & 16.0\stdfmt{1.3} & 31.2\stdfmt{0.9} & 58.2\stdfmt{1.2} & 23.2\stdfmt{1.5} & 4.6\stdfmt{0.8} & 9.8\stdfmt{1.2} & 58.7\stdfmt{1.1} & 23.6\stdfmt{1.2} & 8.5\stdfmt{0.6} & 11.7\stdfmt{1.4} & 55.8\stdfmt{1.0} \\
        \quad w/o Aug. & 23.7\stdfmt{1.2} & 10.1\stdfmt{1.7} & 18.3\stdfmt{1.1} & 34.9\stdfmt{1.5} & 14.0\stdfmt{1.9} & 2.9\stdfmt{1.0} & 6.2\stdfmt{1.6} & 36.4\stdfmt{1.4} & 14.5\stdfmt{1.3} & 5.4\stdfmt{0.8} & 7.4\stdfmt{1.8} & 19.2\stdfmt{1.3} \\
        \midrule
        LLaMA-3 (0-shot) & {37.5\stdfmt{1.1}} & {13.4\stdfmt{0.7}} & {21.3\stdfmt{0.3}} & 42.0\stdfmt{1.2} & {11.3\stdfmt{0.4}} & {2.5\stdfmt{0.2}} & {7.6\stdfmt{1.1}} & 34.7\stdfmt{0.2} & {20.4\stdfmt{1.2}} & {2.3\stdfmt{0.7}} & {9.6\stdfmt{1.3}} & {26.7\stdfmt{1.5}} \\
        LLaMA-3 (1-shot) & 37.8\stdfmt{1.0} & 13.5\stdfmt{0.8} & 21.5\stdfmt{0.4} & 41.7\stdfmt{1.3} & 11.5\stdfmt{0.5} & 2.4\stdfmt{0.2} & 7.8\stdfmt{1.0} & 34.9\stdfmt{0.3} & 20.2\stdfmt{1.1} & 2.4\stdfmt{0.6} & 9.5\stdfmt{1.2} & \textbf{26.9\stdfmt{1.4}} \\
        LLaMA-3 (5-shot) & \highlight{\textbf{44.2\stdfmt{0.9}}} & \highlight{\textbf{19.8\stdfmt{1.1}}} & \highlight{\textbf{36.1\stdfmt{1.2}}} & \highlight{\textbf{64.4\stdfmt{0.7}}} & \highlight{\textbf{26.2\stdfmt{0.6}}} & \highlight{\textbf{5.6\stdfmt{0.4}}} & \highlight{\textbf{12.1\stdfmt{0.6}}} & {\textbf{69.3\stdfmt{1.2}}} & - & - & - & - \\
        \bottomrule
    \end{tabular}
    }
    \caption{\textbf{Summarization Results.} Comparison of different text summarization models on TweetSumm, WikiHow, and ArXiv/PubMed datasets. We report ROUGE-1 (R-1), ROUGE-2 (R-2), ROUGE-L (R-L) F$_1$ scores, and L-Eval scores. We report the mean\stdfmt{std.} performance across 5 different seeds.
    Refer to Appendix~\ref{ssec:impl_details} and Section~\ref{ssec:baselines} for metric and implementation details.
    \textbf{Note.} TSL results are reported from~\citet{zhuang2023self}.
    For EDA and {\method} we use a 50-shot {\dtrain} and generate 1000 examples as {\daug}.
    \textbf{Bold} denotes the best-performing model in a given block and  \highlight{highlight} denotes the \textit{overall} best-performing model.
    For the ArXiv/PubMed dataset, we could fit only 2 documents into LLaMA-3's context (1 from {\dtrain} + 1 generated), so we do not report LLaMA-3 (5-shot).
    }
    \label{tab:full_results}
\end{table*}

\paragraph{L-Eval Scores.} In addition to ROUGE, we use an LLM-based evaluation metric for our task.
Specifically, we use LLaMA-Eval (L-Eval), an open-source variant of the G-Eval metric~\citep{liu-etal-2023-g}, where we prompt {\lmodel} instead of a GPT model.
We use L-Eval as it better aligns with human preferences for text summarization, compared to ROUGE scores and other model-based evaluation metrics, such as BERTScore and BARTScore~\citep{zhang2019bertscore,yuan2021bartscore}.
It is also not biased towards LLM-generated content; however, since LLM-inference speed is low for long documents, we did not compute L-Eval scores during training and only computed them during final testing. 
When computing L-Eval scores, we provide the LLM with the input article and a (generated) summary and instruct it to score the summary on a scale of 1-10 (see Prompt~\ref{prompt:score} in Appendix~\ref{prompt-design} for the full L-Eval prompt template).
Formally, given a test article $A$ and a summary $s$, we compute the L-Eval score as follows:

\begin{equation}
    \textrm{L-Eval}(A, s) = \sum_{r=1}^{10} p_{r} \cdot r,
\end{equation}
where $p_r$ is the probability of generating the rating $r$.
In practice, we can only look at the probabilities of top-5 tokens for {\lmodel}, so we assign a probability of 0 to the remaining ratings (that did not appear in the top-5).

In total, computing \textit{test} L-Eval scores for all the summarization models included in Table~\ref{tab:full_results} took $\sim$5.6 hrs for TweetSumm, $\sim$2.1 days for WikiHow, and $\sim$6 days for the ArXiv/PubMed dataset.

\subsection{Baselines}
\label{ssec:baselines}
We run the following baselines:
\textbf{1) {\method} (Ours).} We augment {\dtrain} using the proposed {\method} approach then train a summarization model on {\dall}.
We also run two variants of this baseline to determine the effect of applying data augmentation and mixup, denoted by \textbf{{\method} w/o Aug.} and \textbf{{\method} w/o Mixup} respectively.
\textbf{2) Easy Data Augmentation (EDA).} We use an edit-based data augmentation technique~\cite{wei-zou-2019-eda} to construct {\daug} instead of using {\method}.
Specifically, we apply the EDA technique to each sentence in an article to construct a new example.
\textbf{3) {\method} (rand.).} Same as \textbf{1)} but {\dtrain} is constructed by randomly selecting $k$ examples from the full training set instead of selecting examples from the $T$ clusters.
We also run \textbf{{\method} (rand.) w/o Aug.} where we do not perform any data augmentation.
\textbf{4) Teacher Student Learning (TSL).} A semi-supervised setup proposed by~\citet{zhuang2023self} that employs a teacher-student learning framework similar to us except they do not use LLM-based pseudo-labeling or relabeling.
We report the performance of the TSL (50:500) and TSL (500:500) models~\footnote{TSL (m:n) denotes access to m labeled examples and n unlabeled examples}.
\textbf{5) {\methodSSL}.} Proposed semi-supervised approach using teacher confidence and prompt-based pseudolabel scoring for text summarization.
We report results for the PPSL (50:250) setting that uses {\lmodel}.
\textbf{6) LLaMA-3-70b ($k$-shot.)} An in-context learning-based approach where we prompt {\lmodel} with $k$ examples randomly selected from  {\dtrain} and then instruct it to summarize a test article.
We use the same prompt as the one we use for summarizing articles (Prompt~\ref{prompt:summarize} in Appendix~\ref{prompt-design}), except we remove the group information and directly populate it with $k$ examples.
\textbf{7) Oracle.} A fully supervised model trained on the complete training set $\mathcal{D}_{train}$ to gauge the upper-bound performance for this task.
\section{Results}
\label{sec:results}
\subsection{{\method} Generates Diverse Documents.}
Table~\ref{tab:qualitative} shows qualitative examples generated by EDA, {\method} w/o mixup and {\method} in Table~\ref{tab:qualitative}.
In the context of Table~\ref{tab:qualitative}, we note that w/o mixup, {\method} generates decent quality documents, but it only covers a single topic (phone/electronic device-related sentences.)
{\method}, on the other hand, generated an example that contains mention of terms from two topics (flight as well as a device-related issue.)
EDA generates the lowest-quality documents with grammatical errors and other artifacts.
However, we note that regardless of the quality of the original document, LLaMA-3-70b generates a high-quality summary in all cases.
\paragraph{Comparison w/ Other DA methods.} From Table~\ref{tab:full_results}, we note that {\method} achieves significantly higher L-Eval and ROUGE scores for both extractive and abstractive summarization tasks.
This demonstrates the superior generation ability of LLMs compared to a simple edit-based DA technique like EDA.
Next, we compare {\method} with {\method} w/o Mixup, a strong LLM-based data augmentation baseline, and note that removing the mixup component from {\method} significantly lowers ROUGE and L-Eval scores across the board (as verified by a T-test).

\subsection{Effect of Clustering Documents.}
We perform a student's T-Test comparing results from {\method} and {\method} (rand.) and note that while ROUGE scores for {\method} are generally higher than {\method} (rand.), the differences are \textit{not} significant.
The only exception was R-2 scores on TweetSumm, where {\method} outperforms {\method} (rand.) by 2.1 points (R-2 of 52.7 v/s 50.6).
On the other hand, the difference in L-Eval scores for the two methods was found to be significant by the T-test for all the datasets.
This further suggests that ROUGE scores might not be able to capture the semantic correctness of the generated summaries and highlights the importance of an LLM-based evaluator that can discern between nuanced semantics in natural language text.
We observe a similar trend after removing the augmentation component from both methods ({\method} w/o Aug. v/s {\method} (rand.) w/o Aug.).

Overall, we conclude that {\method} is better than {\method} (rand.), and we should include diverse examples, if possible, in the prompt as it leads to direct improvements in generation quality.

\subsection{DA v/s SSL Methods}
Comparing {\method} with {\methodSSL} and TSL in Table~\ref{tab:full_results}, we note that our 50-shot {\method} and {\method} (rand.) methods outperform TSL (50:500), which uses 50 labeled examples and 500 unlabeled examples.
Next, our two methods outperform TSL (500:500) on all the metrics except the R-1 score (where the different was not found to be significant).
Overall, {\method} is better than TSL for extractive summarization in extreme data-scarce settings.
Next, we note that {\method} achieves slightly higher ROUGE scores and significantly higher L-Eval scores than PPSL (50:250) for extractive summarization; however, for abstractive summarization, {\method} and {\methodSSL} achieve very similar performance for the three datasets.
Overall, we conclude that prompt-based data augmentation might be better than using a semi-supervised method for extractive summarization in data-scarce setups, but both methods are equally performant for abstractive summarization.

\subsection{Knowledge Distillation from LLaMA-3}
First, we note that increasing the number of examples for the LLaMA-3 method leads to expected improvements in performance except L-Eval scores on the ArXiv/PubMed dataset, where 0-shot and 1-shot LLaMA-3 models achieve similar L-Eval scores.
This may suggest that LLaMA-3 struggles with understanding very long documents.
Next, we note that 0-shot LLaMA-3 outperforms 50-shot {\method} w/o Aug baseline on the TweetSumm dataset in terms of ROUGE scores and L-Eval scores, and it achieves competitive results on ArXiv/PubMed.
Lastly, we note that {\method} achieves competitive performance against LLaMA-3 as a summarizer for both extractive and abstractive tasks, whereas, {\methodSSL} is competitive with LLaMA-3 on only the abstractive task.
Additionally, we note that our methods achieve comparable ROUGE scores to the Oracle model despite using just 50 labels compared to 1000 examples used by the oracle (95\% less).
Overall, we conclude that both {\method} and {\methodSSL} are highly performant models compared to LLaMA-3-70b model, demonstrating effective distillation effect from LLaMA-3-70b to BERT- and DistilBART-based models.
We include additional ablation studies in Appendix~\ref{sec:qualitative} that demonstrate the sample efficiency of {\methodSSL} and show the importance of relabeling and the specific pseudo-labeling strategy used in {\methodSSL}.

\section{Conclusion}
In this work, we focus on low-resource text summarization and propose two novel approaches to effectively employ an LLM for the task: {\method}, a two-step data augmentation method for few-shot summarization, and {\methodSSL}, a multi-step prompt-based semi-supervised framework for sample-efficient semi-supervised text summarization.
Our experiments show that our methods are better than existing approaches for low-resource summarization and that they knowledge transfer from a large teacher model {\lmodel} into much smaller BERT- and DistilBART-based models.
LLM-based approaches are underexplored for low-resource text summarization, and through this work, we hope to spark an interest in the research community to address various challenges of this task.
\section{Limitations}
We use {\lmodel} for our experiments, which has a context window size of 8192 tokens, so it is not possible to fit many long documents in the model's context (like articles in the ArXiv/PubMed dataset).
We can explore using position interpolation (PI) to increase the context window length of LLaMA~\cite{chen2023extending} or switch to more recent LLaMA-3.1 family of models.

Currently, we only consider text summarization for the English language.
Moving forward, we can expand our method to multiple languages.
More research on efficiently handling long documents during the training process is also needed, as we currently rely on a chunk-and-summarize subroutine to train our models for long documents, which results in significant delays in document processing.
We can consider using alternative transformer architectures such as LongFormer~\citep{beltagy2020longformer} as PreSumm's backend.

\section{Ethics Statement}
We generate large textual datasets using LLMs, and even though we use an instruction-tuned model, we need to be careful about any bias it might exhibit, or any potentially harmful content that it might generate.
Language model debiasing is a common potential solution to address this issue~\citep{meade2021empirical,guo2022auto}.
Additionally, we suggest involving a human moderator if these systems are to be made public-facing.
\bibliography{custom}

\appendix

\section{Setting $T$}
\label{sec:T-exp}
We experiment with different values of $T$ (number of groups to divide the training set into) and report the validation performance reported in Table~\ref{tab:setting_T}.
We find that $T=10$ provides the best trade-off between the number of clusters and model performance as increasing $T$ further leads to minimal gains or sometimes no gain at all.

\begin{table}[h!]
    \centering
    \resizebox{0.8\linewidth}{!}
    {%
    \begin{tabular}{lcccc}
        \toprule
        & \multicolumn{2}{c}{\textbf{TweetSumm}} & \multicolumn{2}{c}{\textbf{WikiHow}} \\
        \cmidrule{2-5}
         \textbf{$T$} & \textbf{ROUGE-2} & \textbf{L-Eval} &\textbf{ROUGE-2} & \textbf{L-Eval} \\
        \midrule
        5 & 52.1 & 67.7 & 6.1 & 65.3 \\
        10 & {54.3} & 69.2 & 7.2 & 70.2 \\
        15 & {54.2} & 69.6 & 7.6 & 70.5 \\
        20 & {54.4} & 69.6 & 7.7 & 71.1 \\
        \bottomrule
    \end{tabular}
    }
    \caption{Validation ROUGE-2 and L-Eval scores for different values of $T$ on the TweetSumm datasets.}
    \label{tab:setting_T}
\end{table}

\section{Additional Details for {\methodSSL}.}
\label{sec:ppsl_conf}

\paragraph{Computing Confidence.}
We compute the teacher confidence for a generated summary (a.k.a pseudo-label) as follows: for extractive summarization, and a PreSumm teacher model, let $p_{ij}$ denote the probability with which the $i$-th sentence $s_i$ in an unlabeled document $u_j$ is present in its summary $S_j$, and let $\mathbbm{1}$ denote the indicator function: $\mathbbm{1}(s_i) =
    \begin{cases}
        1, & \text{if}\ s_i \in S \\
        0, & \text{otherwise}
    \end{cases}.$
We then compute the teacher confidence for the pseudo-label $S_j$ by averaging the probabilities of selected sentences.
We define the teacher confidence ($C_j$) for an input text $u_j$ as follows:

\begin{equation}
    C_j = \frac{\sum_{i=1}^{|u_j|} \mathbbm{1}(s_i) \cdot p_{ij}}{n},
\label{eq:conf}
\end{equation}

where $|u_j|$ denotes the number of sentences in the unlabeled document $u_j$ and $n$ is the number of sentences in the generated summary.

\paragraph{Baselines}
We compare our {\methodSSL} with the following baselines:
\textbf{1) PreSumm}~\citep{liu-lapata-2019-text}. The original PreSumm model that pretrains a BERT model for summarization.
We train two PreSumm models -- one on a limited training set with 50 labeled examples to match the starting point of our semi-supervised setting and another with 300 labeled examples, the same as the dataset size at the end of our training cycle.
\textbf{2) Teacher-Student Learning (TSL)}~\citep{zhuang2023self}.
Current state-of-the-art semi-supervised method on TweetSumm.
The teacher-student learning framework uses a similar formulation for computing model confidence to ours, as follows: $C_j = \sum_{i=1}^{n} (C_{ij})/n_j$.
Here, $C_{ij} = p_{ij}q_{ij} + (1 - p_{ij})(1 - q_{ij})$, where $p_{ij}$ is the probability of sentence $i$ being selected for summary for dialog $j$ estimated by the teacher model, and $q_{ij} = 1 \ \textrm{if} \ {p_{ij}} \ \textrm{in top 4, else} \ 0$.
We report the performance of the TSL (50:500) and TSL (500:500) models from the paper, as they are the closest to our setup (50/500 labeled examples + 500 unlabeled examples).
\textbf{3) Confidence + G-4 relabeling + G-4 score (Ours).} Our proposed method following the methodology in Section~\ref{sec:method}.
We first use the PreSumm teacher model to shortlist 50 pseudo-labels (Stage 1 and 2), relabel them using GPT-4 (Stage 3), and then select the top 5 using GPT-4 score (Stage 4).
\textbf{4) Confidence + G-4 score.} We skip Stage 3 from \textbf{3)} to directly score the top 50 PreSumm pseudo-labels using GPT-4.
We run this baseline to measure the effect of relabeling in our pipeline.
\textbf{5) Confidence + G-4 relabeling.} We skip Stage 4 from \textbf{3)} and select the final 5 pseudo-labels based on PreSumm confidence.
\textbf{6) Confidence + L-3 relabeling + L-3 score (Ours).} Same as \textbf{3)} but using LLaMA-3.
\textbf{7) Confidence + L-3 score (Ours).} Same as \textbf{4)} but using LLaMA-3.
\textbf{8) Confidence.} We skip Stage 3 and 4 from from \textbf{3)} and select 5 PreSumm pseudo-labels based on PreSumm confidence.
\textbf{9) Random.} Same as \textbf{6)} but instead of using teacher confidence defined in Equation~\ref{eq:conf}, we randomly select five PreSumm pseudo-labels to include in each cycle.
The results for these baselines is shown in Table~\ref{tab:ppsl_full_results}.

\begin{table*}[t]
    \centering
    \resizebox{\linewidth}{!}
    {%
    \begin{tabular}{lccccccccc}
        \toprule
         & \multicolumn{3}{c}{\textbf{TweetSumm}} & \multicolumn{3}{c}{\textbf{WikiHow}} &
         \multicolumn{3}{c}{\textbf{ArXiv/Pubmed}} \\
         \cmidrule(r){2-10}
         \textbf{Method} & \textbf{R-1 (\%)} & \textbf{R-2 (\%)} & \textbf{R-L (\%)} & \textbf{R-1 (\%)} & \textbf{R-2 (\%)} & \textbf{R-L (\%)} & \textbf{R-1 (\%)} & \textbf{R-2 (\%)} & \textbf{R-L (\%)} \\
        \midrule
        \multicolumn{10}{c}{\textbf{DistilBERT$_{base}$ (50 labels)}}\\
        \midrule
        Random & 36.7 (1.5) & 25.4 (1.4) & 36.7 (1.3) & 19.7 (1.4) & 1.5 (1.1) & 7.2 (1.3) & 19.5 (1.3) & 2.9 (0.9) & 7.8 (1.2) \\
         Confidence & 43.5 (1.4) & 35.1 (1.2) & 46.8 (1.1) & 21.3 (0.4) & 3.7(0.8) & 10.3 (1.0) & 23.4 (1.1) & 5.2 (0.7) & 12.5 (1.1) \\
         \quad + G-4 relabeling & {55.4 (1.3)} & \textbf{46.7 (0.6)} & \underline{56.1 (0.9)} & {22.1 (0.4)} & {5.7 (0.6)} & {13.5 (0.7)} & {23.8 (0.8)} & {7.3 (1.3)} & {15.3 (0.8)} \\
         Confidence + G-4 score & 46.8 (1.3) & 37.4 (0.4) & 48.3 (1.2) & 21.7 (0.9) & 4.6 (0.4) & 12.1 (1.1) & 24.1 (0.9) & 6.7 (0.3) & 13.8 (1.4) \\
        \  + G-4 relabeling (Ours) & \textbf{57.6 (1.2)} & \underline{46.3 (1.7)} & \textbf{56.2 (1.3)} & \textbf{22.7 (0.3)} & \textbf{5.9 (0.4)} & \textbf{13.8 (0.5)} & \textbf{24.7 (0.9)} & \textbf{8.1 (1.3)} & \textbf{15.9 (0.8)} \\
        Confidence + L-3 score & 45.7 (1.1) & 36.9 (0.2) & 47.8 (1.2) & 21.6 (0.4) & 4.1 (0.5) & 11.1 (0.8) & 23.9 (0.9) & 6.1 (0.3) & 12.9 (1.3) \\
        \  + L-3 relabeling (Ours) & \underline{56.2 (1.1)} & {45.1 (1.2)} & {55.9 (1.1)} & \underline{22.3 (0.1)} & \underline{5.8 (0.2)} & \underline{13.6 (0.3)} & \underline{24.5 (0.6)} & \underline{7.7 (1.4)} & \underline{15.7 (0.3)} \\
        \midrule
        \multicolumn{10}{c}{\textbf{BERT$_{base}$ (50 labels)}}\\
        \midrule
        TSL (50:500) & 49.0 & 37.7 & 48.2 & - & - & - & - & - & - \\
        Random & 45.4 (1.4) & 32.4 (1.9) & 42.5 (1.8) & 22.1 (1.7) & 2.4 (1.5) & 9.6 (1.5) & 23.3 (1.4) & 6.1 (1.2) & 12.4 (1.3) \\
         Confidence & 49.7 (1.6) & 39.5 (1.4) & 49.4 (1.3) & 24.5 (0.6) & 4.8 (1.1) & 12.8 (1.0) & 27.6 (1.1) & 7.7 (0.7) & 14.2 (1.2) \\
         \quad + G-4 relabeling & {57.8 (1.2)} & {50.3 (0.5)} & {58.9 (1.2)} & \textbf{26.4 (0.3)} & \textbf{7.3 (0.5)} & \textbf{16.4 (0.8)} & {28.7 (0.9)} & {9.5 (1.1)} & {17.1 (0.8)} \\
         Confidence + G-4 score & 52.3 (1.6) & 42.8 (0.7) & 51.0 (1.4) & 25.2 (0.7) & 5.6 (0.5) & 13.1 (1.0) & 27.7 (0.9) & 7.9 (0.2) & 15.5 (1.3) \\
        \ + G-4 relabeling (Ours) & \textbf{58.9 (1.4)} & \textbf{50.4 (0.8)} & \textbf{59.4 (1.5)} & \underline{26.1 (0.4)} & \underline{7.2 (0.6)} & \underline{15.9 (0.9)} & \textbf{29.1 (0.7)} & \textbf{9.7 (1.2)} & \textbf{17.7 (0.6)} \\
        Confidence + L-3 score & 51.7 (1.2) & 41.6 (1.2) & 50.3 (1.2) & 25.9 (0.3) & 5.2 (0.2) & 13.0 (0.8) & 27.6 (0.4) & 7.9 (0.5) & 15.3 (1.1) \\
        \ + L-3 relabeling (Ours) & \underline{58.4 (1.2)} & {50.1 (0.3)} & \underline{59.1 (1.2)} & {26.0 (0.2)} & {6.9 (0.3)} & {15.1 (0.2)} & \underline{29.0 (0.5)} & \underline{9.4 (0.7)} & \underline{17.4 (0.3)} \\
        \midrule
        \multicolumn{10}{c}{\textbf{BERT$_{base}$ (500 labels)}}\\
        \midrule
        TSL (500:500) & 59.0 & 48.3 & 58.2 & - & - & - & - & - & - \\
        Random & 55.1 (1.4) & 42.7 (1.1) & 50.3 (1.2) & 25.6 (1.3) & 4.5 (1.1) & 15.2 (1.3) & 25.4 (1.5) & 9.5 (1.2) & 24.1 (1.2) \\
         Confidence & 61.8 (0.7) & 54.9 (0.8) & 60.3 (0.9) & 28.4 (0.6) & 8.0 (1.1) & 22.5 (1.0) & 29.4 (0.5) & 11.5 (0.6) & 27.7 (0.8) \\
        \ + L-3 score & 63.4 (0.5) & 55.6 (0.8) & 62.1 (0.7) & 28.9 (0.4) & 8.2 (0.5) & 28.4 (0.4) & 31.7 (0.3) & 11.8 (0.2) & 29.4 (0.4) \\
        \ + L-3 relabeling (Ours) & \textbf{64.2 (0.2)} & \textbf{56.2 (0.4)} & \textbf{62.8 (0.6)} & \textbf{30.7 (0.4)} & \textbf{8.8 (0.3)} & \textbf{29.5 (0.3)} & \textbf{33.5 (0.3)} & \textbf{12.3 (0.2)} & \textbf{32.2 (0.3)} \\
        \bottomrule
    \end{tabular}
    }
    \caption{Mean (Std.) ROUGE F-1 scores of different pseudo-labeling strategies.
    R-1, R-2, and R-L denote ROUGE-1, ROUGE-2, and ROUGE-L metrics, respectively.
    TSL results from~\cite{zhuang2023self}.
    Refer to Section~\ref{sec:exp_setup} for method details.
    \textbf{Bold} indicates the best-performing and  \underline{underline} denotes the second-best performing method, respectively.}
    \label{tab:ppsl_full_results}
\end{table*}

\begin{table}[t!]
    \centering
    \resizebox{\linewidth}{!}
    {%
    \begin{tabular}{lcccccc}
        \toprule
         & \multicolumn{2}{c}{\textbf{TweetSumm}} & \multicolumn{2}{c}{\textbf{WikiHow}} &
         \multicolumn{2}{c}{\textbf{ArXiv/Pubmed}} \\
         \cmidrule(r){2-7}
         \textbf{Method} & \textbf{R-2} & \textbf{L-Eval} & \textbf{R-2} & \textbf{L-Eval}  & \textbf{R-2} & \textbf{L-Eval} \\
        \midrule
        PreSumm (50 labels) & 37.1 (1.1) & 31.2 (0.5) & 3.2 (0.8) & 34.2 (1.5) & 7.3 (0.9) & 13.5 (1.2) \\
        PreSumm (300 labels) & 51.1 (2.1) & {60.5 (1.2)} & 7.6 (0.6) & {68.1 (1.1)} & 10.8 (0.9) & {49.5 (2.4)} \\
        PreSumm (500 labels) & 54.4 (1.2) & {67.1 (0.3)} & 7.9 (0.5) & {74.4 (0.6)} & 11.3 (0.5) & {58.2 (1.1)} \\
        PreSumm (750 labels) & \underline{56.1 (0.7)} & \underline{70.3 (0.5)} & \underline{8.5 (0.4)} & \underline{76.5 (0.4)} & \underline{12.1 (0.7)} & \underline{62.8 (0.7)} \\
        \midrule
        \multicolumn{7}{c}{\textbf{50 labels}} \\
        \midrule
        Random & 32.4 (1.9) & 32.1 (1.1) & 2.4 (1.5) & 37.7 (1.6) & 6.1 (0.2) & 15.1 (2.3) \\
         Confidence + G-4 score & 42.8 (0.7) & 46.2 (0.2) & 5.6 (0.5) & 59.4 (1.3) & 7.9 (0.2) & 40.1 (1.9) \\
        \ + G-4 relabeling (Ours) & {50.4 (0.8)} & {58.4 (0.4)} & {7.2 (0.6)} & {70.3 (1.4)} & {9.7 (1.2)} & {52.5 (1.3)} \\
        Confidence + L-3 score & 41.6 (1.2) & 45.8 (0.7) & 5.2 (0.2) & 57.5 (1.4) & 7.9 (0.5) & 37.1 (1.8) \\
        \ + L-3 relabeling (Ours) & {50.1 (0.3)} & {56.3 (0.9)} & {6.9 (0.3)} & {69.3 (2.1)} & {9.4 (0.7)} & {49.3 (1.4)} \\
        \midrule
        \multicolumn{7}{c}{\textbf{500 labels}} \\
        \midrule
        Random & 42.7 (1.1) & 52.3 (1.2) & 4.5 (1.1) & 52.7 (1.8) & 9.5 (1.2) & 44.1 (0.9) \\
        Confidence + L-3 score & 55.6 (0.8) & 69.2 (0.7) & 8.2 (0.5) & 75.2 (1.4) & 11.8 (0.2) & 60.2 (0.5) \\
        \ + L-3 relabeling (Ours) & \textbf{56.2 (0.8)} & \textbf{71.2 (0.9)} & \textbf{8.8 (0.3)} & \textbf{77.3 (1.3)} & \textbf{12.3 (0.2)} & \textbf{65.7 (0.3)} \\
        \bottomrule
    \end{tabular}
    }
    \caption{\textbf{Fully-supervised methods (first four rows) semi-supervised approaches (remaining rows).} All models use BERT$_{base}$ as PreSumm's backbone. The number of labeled examples for fully supervised models is shown in brackets.
    The semi-supervised methods use 50/500 labeled and 250 unlabeled examples.}
    \label{tab:sup_vs_semisup}
\end{table}

\section{Implementation Details}
\label{ssec:impl_details}
\textbf{Data Augmentation.} We set the number of groups $T=10$ for all datasets\footnote{based on validation R-2 and L-Eval scores reported in Table~\ref{tab:setting_T} of Appendix~\ref{sec:T-exp}} and randomly sample 5 examples from each group to get a $50$-shot {\dtrain}.
Then, we obtain {\daug} by generating 1000 examples using the procedure described in Section~\ref{sec:method}.
In the data generation prompt, we include five examples for each group for TweetSumm and WikiHow, but for ArXiv/PubMed, we could only fit two documents at a time in LLaMA-3's context window after applying the following truncation heuristic to the text.
We include $l$ lines before and after each sentence in the ground truth summary such that we are able to fit two examples in the prompt.
The average value of $l$ was 5.21 (so approximately $\sim$90 for an average summary size of 8 sentences for the ArXiv/PubMed dataset sentences were selected for example\footnote{$(5.21 \times 2 \times 8) + 8 = 91.1$}).
Here, we set the summary size $p$ to 4 sentences for TweetSumm and WikiHow datasets, and 8 sentences for the Arxiv/PubMed dataset.
We determine these summary sizes based on the average summary size in the few-shot training data {\dtrain}.
We host {\lmodel} on 4$\times$A100 GPUs with 80G VRAM each and use it as the backbone LLM for all our experiments.
Generating {\daug} took $\sim$4.2 hrs for TweetSumm, $\sim$11.3 hrs for WikiHow, and $\sim$1.4 days for ArXiv/PubMed dataset.

\noindent \textbf{Training.} For extractive summarization, we train a PreSumm model on the combined {\method}-generated and seed few-shot dataset {\dall}.
We use the TransformerSum repository\footnote{\url{https://transformersum.readthedocs.io/}} to implement our training pipeline.
To handle long documents that cannot be fed to the PreSumm at once, we introduce a subroutine that iteratively chunks and summarizes the document until we obtain a summary of size $p$.
The iterative subroutine is crucial to train PreSumm models on the WikiHow and ArXiv/PubMed datasets with long input documents.
For abstractive summarization, we follow an \textit{extractive-then-abstractive} approach, where for a given input document, we first obtain its extractive summary using the full-trained PreSumm model from the previous step.
Then, we finetune a DistilBART model that summarizes the PreSumm summaries to generate abstractive summaries.

We initialize the training process with a learning rate of $2 \times 10^{-5}$ and use a cyclic learning rate scheduler~\citep{smith2015cyclical}.
We train all our models for 100 epochs with an early stopping criterion, where we stop the training process if the validation ROUGE-2 score does not improve for more than 10 epochs.
We use the AdamW optimizer~\cite{loshchilov2017decoupled} with $\epsilon = 1 \times 10^{-8}, \beta_1 = 0.9, \beta_2 = 0.99$ and train all our models on one V100 GPU with 12G VRAM.
We use \texttt{distilbart-12-6-cnn} backbone for abstractive summarization and experiment with two backbones for the PreSumm model: DistilBERT$_{base}$ and BERT$_{base}$ (results in Table~\ref{tab:full_res}) and find BERT$_{base}$ to be better.
Training a model on {\method}-generated data took $\sim$2.5 hrs for TweetSumm, $\sim$13.4 hrs, for WikiHow, and $\sim$2.7 days for ArXiv/PubMed.
Crucially, we repeat each experiment (data augmentation+model training) for 5 random seeds and report the mean and standard deviations for all models unless otherwise stated.

\begin{table*}[t]
    \centering
    \resizebox{\linewidth}{!}
    {%
    \begin{tabular}{lcccccccccccc}
        \toprule
         & \multicolumn{4}{c}{\textbf{TweetSumm}} & \multicolumn{4}{c}{\textbf{WikiHow}} &
         \multicolumn{4}{c}{\textbf{ArXiv/Pubmed}} \\
         \cmidrule(r){2-13}
         \textbf{Method} & \textbf{R-1 (\%)} & \textbf{R-2 (\%)} & \textbf{R-L (\%)} & \textbf{L-Eval (\%)} & \textbf{R-1 (\%)} & \textbf{R-2 (\%)} & \textbf{R-L (\%)} & \textbf{L-Eval (\%)} & \textbf{R-1 (\%)} & \textbf{R-2 (\%)} & \textbf{R-L (\%)} & \textbf{L-Eval (\%)} \\
        \toprule
        \multicolumn{13}{c}{\textbf{DistilBERT$_{base}$}}\\
        \midrule
        \textit{Oracle} & \textit{62.8\stdfmt{0.6}} & \textit{53.1\stdfmt{1.2}} & \textit{59.3\stdfmt{0.7}} & \textit{83.6\stdfmt{0.5}} & \textit{30.7\stdfmt{0.4}} & \textit{8.6\stdfmt{0.8}} & \textit{19.1\stdfmt{0.7}} &\textit{81.6\stdfmt{1.2}} & \textit{34.2\stdfmt{0.6}} & \textit{12.3\stdfmt{1.2}} & \textit{19.4\stdfmt{0.4}} & \textit{71.1\stdfmt{0.4}} \\
        \toprule
        PPSL (50:250) & {56.2\stdfmt{1.1}} & {45.1\stdfmt{1.2}} & {55.9\stdfmt{1.1}} & - & {22.3\stdfmt{0.1}} & {5.8\stdfmt{0.2}} & {13.6\stdfmt{0.3}} & - & {24.5\stdfmt{0.6}} & {7.7\stdfmt{1.4}} & {15.7\stdfmt{0.3}} & - \\
        EDA & {47.3\stdfmt{1.3}} & {36.1\stdfmt{1.2}} & {48.7\stdfmt{1.2}} & 51.3\stdfmt{0.3} & {21.6\stdfmt{0.8}} & {3.3\stdfmt{0.8}} & {11.8\stdfmt{1.2}} & 54.1\stdfmt{0.3} & {23.3\stdfmt{1.3}} & {5.4\stdfmt{0.6}} & {12.6\stdfmt{1.3}} & 39.6\stdfmt{0.2} \\
        {\method} (rand.) & {56.9\stdfmt{2.5}} & {46.1\stdfmt{3.4}} & {58.7\stdfmt{3.1}} & 56.7\stdfmt{0.6} & {22.7\stdfmt{2.1}} & {6.1\stdfmt{1.2}} & {14.8\stdfmt{1.3}} & 65.9\stdfmt{0.4} & {24.8\stdfmt{1.5}} & {8.3\stdfmt{1.7}} & {16.0\stdfmt{1.3}} & 48.1\stdfmt{0.6} \\
        \quad w/o Aug. & {41.7\stdfmt{1.6}} & {32.4\stdfmt{1.2}} & {43.6\stdfmt{2.1}} & 23.4\stdfmt{1.2} & {19.2\stdfmt{1.8}} & {2.1\stdfmt{0.6}} & {9.1\stdfmt{1.4}} & 20.4\stdfmt{1.0} & {21.4\stdfmt{1.2}} & {4.7\stdfmt{0.3}} & {10.4\stdfmt{1.2}} & 13.3\stdfmt{0.5} \\
        {\method} (ours) & \textbf{57.3\stdfmt{2.4}} & \textbf{46.8\stdfmt{3.1}} & \textbf{57.2\stdfmt{2.7}} & \textbf{60.3\stdfmt{0.5}} & \textbf{23.4\stdfmt{1.7}} & \textbf{6.5\stdfmt{1.6}} & \textbf{15.2\stdfmt{1.1}} & \textbf{68.4\stdfmt{1.3}} & \textbf{25.7\stdfmt{1.7}} & \textbf{8.6\stdfmt{2.1}} & \textbf{16.6\stdfmt{1.4}} & \textbf{51.2\stdfmt{0.6}} \\
        \quad w/o Mixup & {54.2\stdfmt{1.7}} & {44.3\stdfmt{1.4}} & {53.5\stdfmt{1.4}} & 55.3\stdfmt{1.2} & {22.1\stdfmt{1.3}} & {4.7\stdfmt{0.2}} & {12.8\stdfmt{1.2}} & 62.3\stdfmt{0.7} & {23.8\stdfmt{1.2}} & {6.1\stdfmt{0.9}} & {14.1\stdfmt{1.3}} & 42.1\stdfmt{1.1} \\
        \quad w/o Aug. & {42.8\stdfmt{1.1}} & {34.1\stdfmt{1.1}} & {44.2\stdfmt{1.4}} & 28.4\stdfmt{0.8} & {19.7\stdfmt{1.2}} & {2.8\stdfmt{0.4}} & {10.2\stdfmt{1.1}} & 31.4\stdfmt{0.4} & {22.6\stdfmt{1.3}} & {4.9\stdfmt{0.6}} & {11.3\stdfmt{1.3}} & 18.6\stdfmt{0.5} \\
        \toprule
        \multicolumn{13}{c}{\textbf{BERT$_{base}$}}\\
        \midrule
        \textit{Oracle} & \textit{65.7\stdfmt{0.3}} & \textit{56.6\stdfmt{0.4}} & \textit{64.9\stdfmt{0.4}} & \textit{86.2\stdfmt{0.3}} & \textit{30.5\stdfmt{0.5}} & \textit{8.7\stdfmt{0.3}} & \textit{19.2\stdfmt{0.6}} & \textit{87.3\stdfmt{0.1}} & \textit{34.6\stdfmt{0.4}} & \textit{12.4\stdfmt{0.2}} & \textit{19.6\stdfmt{0.4}} & \textit{78.1\stdfmt{0.5}} \\
        \toprule
        TSL (50:500) & 49.0 & 37.7 & 48.2 & - & - & - & - & -  & - & - & - & - \\
        TSL (500:500) & 59.0 & 48.3 & 58.2 & - & - & - & - & - & - & - & - & - \\
        EDA & {51.1\stdfmt{0.7}} & {39.2\stdfmt{0.9}} & {53.0\stdfmt{0.2}} & 34.3\stdfmt{1.2} & {23.4\stdfmt{0.5}} & {4.1\stdfmt{0.3}} & {13.0\stdfmt{0.5}} & 42.1\stdfmt{0.8} & {26.2\stdfmt{1.1}} & {7.9\stdfmt{1.0}} & {13.1\stdfmt{0.6}} & 17.2\stdfmt{0.5} \\
        PPSL (50:250) & {58.4\stdfmt{1.2}} & {50.1\stdfmt{0.3}} & {59.1\stdfmt{1.2}} & 56.3\stdfmt{0.9} & {26.0\stdfmt{0.2}} & {6.9\stdfmt{0.3}} & {15.1\stdfmt{0.2}} & 69.3\stdfmt{2.1} & {29.0\stdfmt{0.5}} & {9.4\stdfmt{0.7}} & {17.4\stdfmt{0.3}} & 49.3\stdfmt{1.4} \\
        {\method} (rand.) & {58.6\stdfmt{3.2}} & {50.6\stdfmt{2.1}} & {59.7\stdfmt{2.3}} & 60.3\stdfmt{0.9} & {26.4\stdfmt{1.0}} & {7.5\stdfmt{1.2}} & {15.8\stdfmt{0.2}} & 72.5\stdfmt{1.2} & {30.7\stdfmt{1.7}} & \highlight{\textbf{10.6\stdfmt{1.5}}} & \highlight{\textbf{18.5\stdfmt{1.1}}} & 48.4\stdfmt{1.1} \\
        \quad w/o Aug. & {49.4\stdfmt{0.7}} & {36.9\stdfmt{1.0}} & {49.0\stdfmt{0.2}} & 31.5\stdfmt{0.5} & {21.3\stdfmt{0.4}} & {3.2\stdfmt{0.4}} & {11.4\stdfmt{0.5}} & 34.2\stdfmt{1.5} & {23.4\stdfmt{1.1}} & {7.5\stdfmt{1.4}} & {12.3\stdfmt{0.8}} & 13.5\stdfmt{1.2} \\
        {\method} (ours) & \textbf{59.1\stdfmt{1.7}} & \textbf{52.7\stdfmt{1.6}} & \highlight{\textbf{60.5\stdfmt{1.3}}} & \textbf{65.3\stdfmt{1.2}} & \textbf{27.3\stdfmt{2.1}} & \highlight{\textbf{7.8\stdfmt{1.3}}} & \textbf{16.6\stdfmt{1.8}} & \highlight{\textbf{81.1\stdfmt{1.7}}} & {\textbf{31.2\stdfmt{1.2}}} & \highlight{\textbf{10.7\stdfmt{1.1}}} & \highlight{\textbf{18.3\stdfmt{1.1}}} & \highlight{\textbf{53.1\stdfmt{0.5}}} \\
        \quad w/o Mixup & {56.1\stdfmt{1.1}} & {47.3\stdfmt{1.2}} & {55.3\stdfmt{1.1}} & 57.3\stdfmt{0.5} & {25.7\stdfmt{1.4}} & {6.2\stdfmt{1.2}} & {14.7\stdfmt{0.7}} & 67.3\stdfmt{2.1} & {28.4\stdfmt{1.9}} & {8.3\stdfmt{1.3}} & {16.8\stdfmt{1.6}} & 52.3\stdfmt{1.2} \\
        \quad w/o Aug. & {50.1\stdfmt{0.6}} & {38.1\stdfmt{1.0}} & {49.9\stdfmt{0.6}} & 32.3\stdfmt{3.1} & {21.9\stdfmt{0.3}} & {3.5\stdfmt{0.2}} & {12.1\stdfmt{0.9}} & 33.3\stdfmt{1.7} & {24.1\stdfmt{0.9}} & {7.9\stdfmt{1.0}} & {12.7\stdfmt{0.5}} & 19.0\stdfmt{2.5} \\
        \midrule
        LLaMA-3 (0-shot) & {50.3\stdfmt{0.5}} & {47.7\stdfmt{0.4}} & {49.9\stdfmt{0.3}} & 52.3\stdfmt{1.2} & {12.2\stdfmt{0.2}} & {2.7\stdfmt{0.5}} & {8.1\stdfmt{0.4}} & 32.3\stdfmt{0.3} & {23.6\stdfmt{0.2}} & {4.6\stdfmt{0.7}} & {15.4\stdfmt{0.3}} & \textbf{38.4\stdfmt{0.5}} \\
        LLaMA-3 (1-shot) & {51.7\stdfmt{0.2}} & {49.2\stdfmt{0.3}} & {51.9\stdfmt{0.3}} & 58.7\stdfmt{1.1} & {14.3\stdfmt{0.2}} & {4.1\stdfmt{0.5}} & {10.6\stdfmt{0.2}} & 39.4\stdfmt{0.5} & \highlight{\textbf{32.6\stdfmt{0.4}}} & \textbf{6.5\stdfmt{0.7}} & \textbf{17.2\stdfmt{0.3}} & 38.3\stdfmt{1.8} \\
        LLaMA-3 (5-shot) & \highlight{\textbf{62.4\stdfmt{0.5}}} & \highlight{\textbf{54.3\stdfmt{0.7}}} & \highlight{\textbf{60.3\stdfmt{1.1}}} & \highlight{\textbf{67.5\stdfmt{0.6}}} & \highlight{\textbf{28.7\stdfmt{0.3}}} & {\textbf{7.5\stdfmt{0.9}}} & \highlight{\textbf{17.1\stdfmt{0.3}}} & {\textbf{71.3\stdfmt{0.4}}} & - & - & - & - \\
        \bottomrule
    \end{tabular}
    }
    \caption{\textbf{Extractive Summarization Results.} Comparison of different text summarization models on TweetSumm, WikiHow, and ArXiv/PubMed datasets. We report ROUGE-1 (R-1), ROUGE-2 (R-2), ROUGE-L (R-L) F$_1$ scores, and L-Eval scores. We report the mean\stdfmt{std.} performance across 5 different seeds.
    Refer to Appendix~\ref{ssec:impl_details} and Section~\ref{ssec:baselines} for metric and implementation details.
    \textbf{Note.} TSL results are reported from~\citet{zhuang2023self}.
    For EDA and {\method} we use a 50-shot {\dtrain} and generate 1000 examples as {\daug}.
    \textbf{Bold} denotes the best-performing model in a given block and  \highlight{highlight} denotes the \textit{overall} best-performing model.
    For the ArXiv/PubMed dataset, we could fit only 2 documents into LLaMA-3's context (1 from {\dtrain} + 1 generated), so we do not report LLaMA-3 (5-shot).
    }
    \label{tab:full_res}
\end{table*}

\paragraph{Semi-Supervised Text Summarization.}
We use the TransformerSum repository\footnote{\url{https://transformersum.readthedocs.io/en/latest/}} to implement our training pipeline.
We use PreSumm as our teacher model $M$ and experiment with two backbones: \texttt{distilbert-base-uncased} and \texttt{bert-base-uncased}.
We perform experiments in two settings: \textbf{1)} data-scarce setting where fix the size of the labeled set $D_l$ to 50 for all the datasets, and \textbf{2)} data-abundant setting where we set the size of $D_l$ to 500.
We set $N_{cycles}$ to 50 for all experiments.
We add 5 pseudo-labels to the training set in each cycle, thus resulting in a final training set size of 300 (50 labeled + 250 pseudo-labeled examples.)
We set the summary size $k$ to 4 for TweetSumm and 8 for WikiHow and ArXiv/PubMed.
We base these summary sizes on the average summary size of the labeled training set.
For training, we start with a learning rate of $2 \times 10^{-5}$ on all the datasets and use a cyclic learning rate scheduler during training~\citep{smith2015cyclical}, which is the default setting in TransformerSum.
Additionally, we use AdamW as our optimizer with $\epsilon = 1 \times 10^{-8}, \beta_1 = 0.9, \beta_2 = 0.99$.
We train all our models on a single V100 GPU with 12G VRAM.
We repeat each experiment for three different seeds and report the mean and standard deviation in our results unless otherwise stated.

  
  


\section{Qualitative Results}
\label{sec:qualitative}
\begin{table*}[htbp]
\centering
\scriptsize
\begin{tabular*}{\textwidth}{p{6.1in}}
\toprule
\multicolumn{1}{c}{\textbf{EDA}} \\
\toprule
\textbf{Generated Document:}
\yellow{Customer:@VerizonSupport} I penury need help. I do not have your service (6+ iodine hours to setup and cancel an install!) but the premature homeowner did. There is an ONT in the basement that every out a loud, very annoying beep sends 15 minutes? This is ridiculous. How do I iodine stop it?
Agent:@633638 Press and hold the alarm silence button/ONT reboot button on the bring up unit for 30 seconds. That bequeath will silence it.
Customer:@VerizonSupport done. Will reply back respond later if it beeps again. @VerizonSupport comprise I expected to do this every 24 hours? That's how long the Alarm Silence lasted. Bedden back to beeping every 15 moment since this morning.
Agent:@633638 assay unplugging the power and removing the battery. indium After 3mins, put the battery back in and power on. This should assistant with the beeping. Let the states know. \^DDG
Customer:@VerizonSupport I no service with you at all. Can't I just unplug it and remove assault and \yellow{battery the battery forever?}
Agent:@633638 Yes, you can do that since you do not inspection and repair have service with us. personify When the battery is dead the beeping will stop. \^HSB
Customer:@VerizonSupport you! Thank
Agent:@633638 personify You are welcome! \\
\midrule
\textbf{LLaMA-3 Summary:}
Customer:@VerizonSupport I penury need help. There is an ONT in the basement that every out a loud, very annoying beep sends 15 minutes? This is ridiculous. How do I iodine stop it?
Agent:@633638 Press and hold the alarm silence button/ONT reboot button on the bring up unit for 30 seconds. That bequeath will silence it.
Customer:@VerizonSupport done. Will reply back respond later if it beeps again. @VerizonSupport comprise I expected to do this every 24 hours?
Agent:@633638 assay unplugging the power and removing the battery. \\
\toprule
\multicolumn{1}{c}{\textbf{{\method} w/o mixup}} \\
\toprule
\textbf{Generated Document:}
Customer: \yellow{My iPhone's battery} is draining much faster than usual, and it's overheating. Any suggestions? Here's a link to a screenshot of my battery usage: https://t.co/xyzABC123
@AppleSupport can you help with this issue?
Agent: @UserHelp123 We're here to assist you.
First, could you let us know which iOS version your iPhone is currently using?
Customer: \yellow{@AppleSupport} It's updated to iOS 14.4.
Agent: @UserHelp123 Thanks for the information.
 Have you installed any new apps or updates around the time you started noticing these issues?
Customer: @AppleSupport Yes, I installed a few new games last week.
Agent: @UserHelp123 It's possible that those games are causing the battery drain and overheating.
Could you try removing them to see if the situation improves?
Customer: @AppleSupport I've uninstalled the games, but my phone is still overheating.
Agent: @UserHelp123 Understood.
 Let's perform a \yellow{soft reset on your iPhone.} This won't erase your data but can help with such issues.
 If the problem persists, we should continue this conversation in DM to get more details and possibly set up a service appointment.
 Please reach out in DM with the exact time you started experiencing these issues and if you've noticed any specific patterns.
 https://t.co/NmDpIWxYfO \\
 \midrule
 \textbf{LLaMA-3 Summary:}
 Customer: My iPhone's battery is draining much faster than usual, and it's overheating. Any suggestions?
Agent: Have you installed any new apps or updates around the time you started noticing these issues?
Customer: @AppleSupport I've uninstalled the games, but my phone is still overheating.
Agent: Let's perform a soft reset on your iPhone. This won't erase your data but can help with such issues. \\
\toprule
\multicolumn{1}{c}{\textbf{{\method}}} \\
\toprule
\textbf{Generated Document:}
\yellow{Customer:@VerizonSupport} I recently moved into a house that has your equipment but I don't use your service. The device in the basement beeps every 20 minutes. How do I stop it?
Agent:@VerizonHelper Please press and \yellow{hold the alarm silence button} on the ONT for about 30 seconds. That should silence it temporarily.
Customer:@VerizonSupport I did that, but it started beeping again after a few hours. I need a permanent fix. \cyan{I have a flight next week, and I can't deal with this noise.}
Agent:@VerizonHelper If the alarm silence is temporary, try unplugging the device and removing its battery. After a few minutes, reconnect the battery and power it on.
Customer:@VerizonSupport \cyan{I need this sorted out before I leave.} Is there no faster way?
Agent:@VerizonHelper For a quick solution, you can permanently disconnect the power and \yellow{remove the battery.} This will stop the beeping entirely.
Customer:@VerizonSupport Thank you for the advice! I'll handle it immediately. \\
\midrule
\textbf{LLaMA-3 Summary:}
Customer:@VerizonSupport I recently moved into a house that has your equipment but I don't use your service. The a device in the basement beeps every 20 minutes. How do I stop it?
Agent:@VerizonHelper Please press and hold the alarm silence button on the ONT for about 30 seconds. That should silence it temporarily.
Customer:@VerizonSupport I did that, but it started beeping again after a few hours. I need a permanent fix. I have a flight next week, and I can't deal with this noise.
Agent:@VerizonHelper For a quick solution, you can permanently disconnect the power and remove the battery. This will stop the beeping entirely. \\
\bottomrule
\end{tabular*}
\caption{Full qualitative comparison of EDA, {\method} w/o mixup, and {\method} examples. For each method, the first row shows the generated document and the second row shows its LLaMA-3-generated summary. For these examples, group 1 contained customer conversations with phone companies, such as Verizon, and group 2 contained customer interactions with airline representatives, such as AirAsia and Delta.  \textbf{Note.} For brevity, we do not include the input examples used in the prompt, and for EDA, we perform augmentations one sentence in the document at a time. We note that w/o mixup, {\method} generated a decent quality document but it only covers phone/electronic device-related sentences. {\method}, on the other hand, generated an example that contains mention of flight as well as a device-related issue. Additionally, we note that no matter the quality of the document, LLaMA-3-70b generates a high-quality summary in all cases.}
\label{tab:qualitative}
\end{table*}

\begin{figure*}[t]
    \centering
    \includegraphics[width=\linewidth]{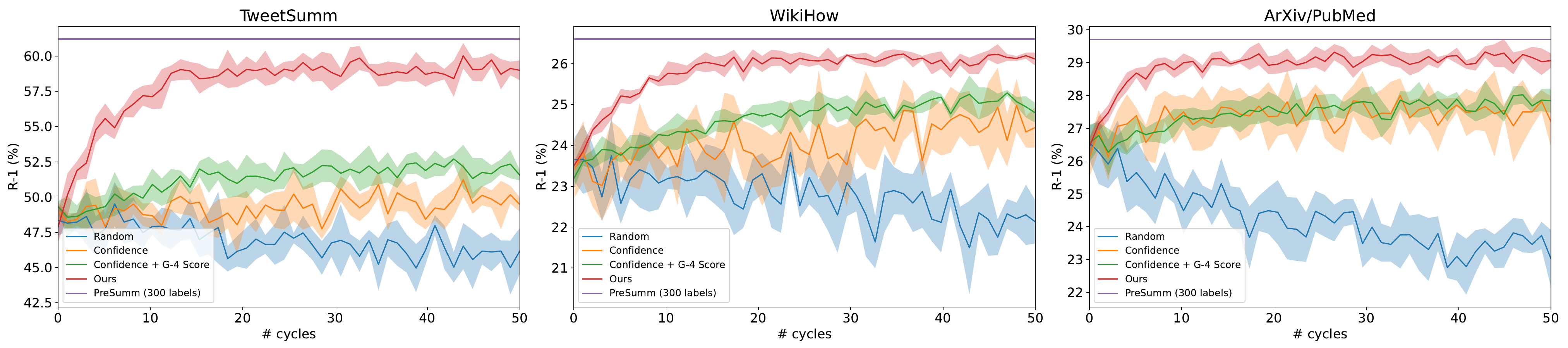}
    \caption{\textbf{ROUGE-1 curves v/s \# cycles for data-scarce setting.} Each cycle denotes an addition of 5 new pseudo-labels to the training set. All results use BERT$_{base}$ as the backbone for PreSumm. The curves are averaged for three seeds (the width denotes the std). Note that we report the GPT-4 version of our method here.}
    \label{fig:r2_graph}
\end{figure*}

\subsection{On the Sample Efficiency of {\methodSSL}}
We now compare the sample efficiency of {\methodSSL} against other methods.
Referring to Table~\ref{tab:sup_vs_semisup},
for fully supervised methods, we note that including more labeled examples improves L-Eval and ROUGE scores across the board (``PreSumm (50 labels)" v/s ``PreSumm (300 labels)").
Our semi-supervised approach using 50 labels with GPT-4 relabeling and GPT-4 score achieves competitive performance to the fully supervised PreSumm model trained on 300 labels.
Notably, we get better L-Eval scores than ``PreSumm (300 labels)" on WikiHow and ArXiv/PubMed datasets and are competitive on TweetSumm.
Note that the ``PreSumm (300 labels)" model approximates the best-case scenario when all the labels in the training set are high-quality.
This is encouraging, as ``PreSumm (300 labels)" approximates the best-case scenario of 100\% high-quality labels in the training set.
In the data-abundant setting, our proposed method with LLaMA outperforms the respective fully supervised model in terms of both ROUGE and L-Eval.
From Table~\ref{tab:full_results}, we further note that our approach outperforms TSL (50:500) while using half the number of pseudo-labels.
We may further improve the model performance by including some examples in the prompt.
Our proposed method outperforms TSL (50:500) and TSL (500:500) despite working in a more challenging labeled:unlabeled dataset ratio of 50:250.
We plot the R-1 scores against the number of training cycles for {\methodSSL} and other semi-supervised baselines (refer to Section~\ref{sec:ppsl_conf} in Appendix~\ref{sec:ppsl_conf} for more details) in
Figure~\ref{fig:r2_graph}. 
Overall, ``Random" setting is highly unstable, ``Confidence + G-4 score" slightly improves over ``Confidence" on TweetSumm and WikiHow, but more importantly, it is consistently more stable.
Finally, our method with GPT-4 scoring and relabeling not only significantly boosts the R-1 scores (visible gap between ``Ours" and the rest) but also does so at a much faster rate.
For all the datasets, our method peaks and stabilizes under 20 cycles (100 pseudo-labels), further endorsing the sample efficiency of our method compared to other approaches.
\subsection{Comparison of Pseudo-label Selection Strategies}
\label{sec:pseudo-quality}
Referring to Table~\ref{tab:full_results}, we note that all pseudo-label selection strategies outperform the random baseline.
The ``Random" baseline performs worse than the fully supervised counterpart on all datasets (R-2 in Table~\ref{tab:full_results} v/s R-2 in Table~\ref{tab:sup_vs_semisup}), meaning that the \textit{majority} of the shortlisted PreSumm pseudo-labels are low-quality.
Using teacher confidence leads to slight performance gains on all the datasets, and adding GPT-4 score further improves the results (``Confidence" v/s ``Confidence + G-4 score" in Table~\ref{tab:full_results}).
These improvements indicate that the shortlisted PreSumm pseudo-labels include some good-quality pseudo-labels, too, and using GPT-4 to rate those pseudo-labels is crucial to picking them.
We see similar trends when using LLaMA-3.

\begin{figure}[htbp]
    \centering
    \includegraphics[width=\linewidth]{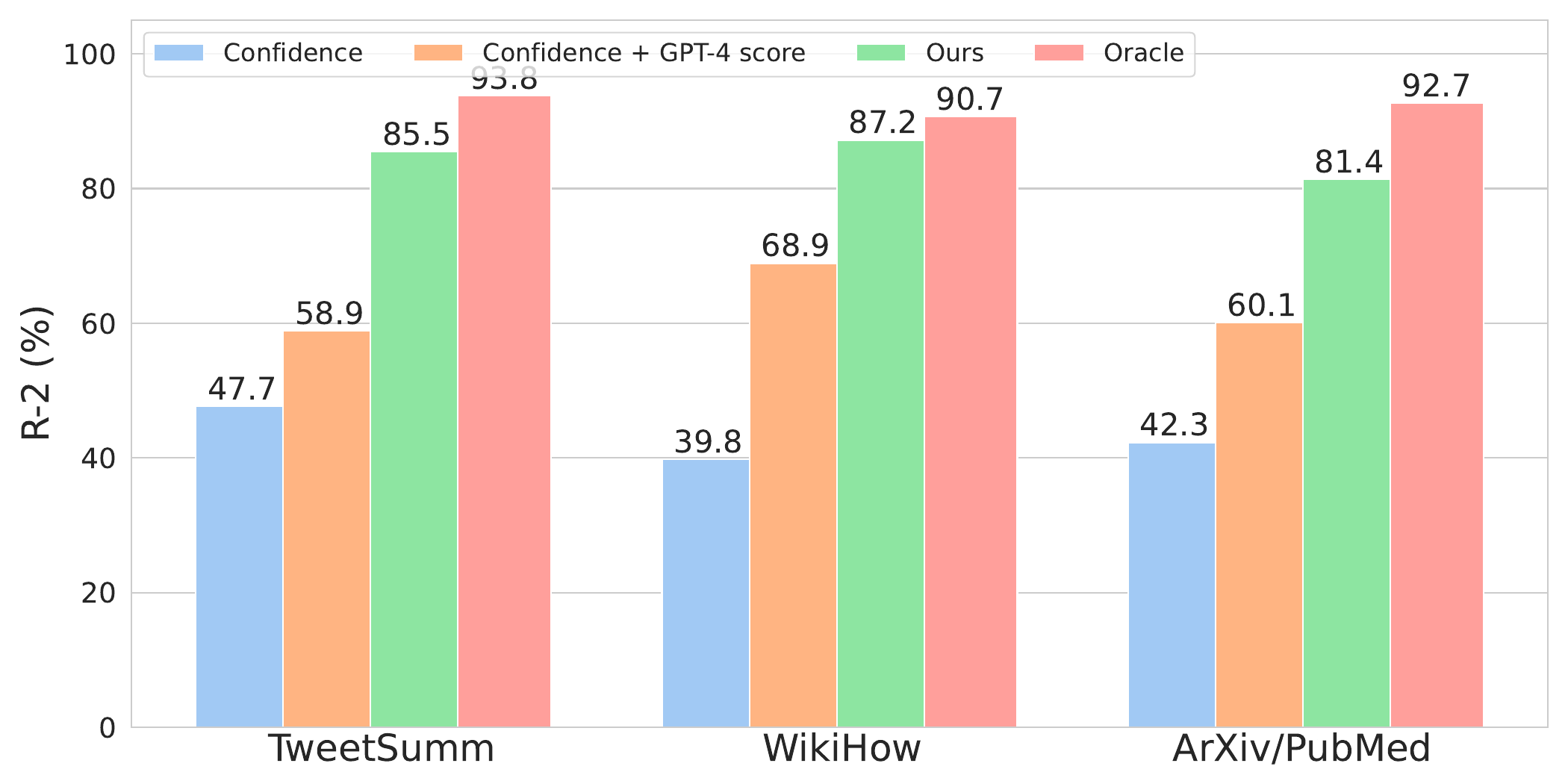}  
    \caption{\textbf{Quality of pseudo-labels by different strategies (data-scarce setup).} The y-axis denotes the ROUGE-2 scores of the top 5 pseudo-labels computed against the respective ground truths. All results are for BERT$_{base}$ as the backbone for PreSumm and three random seeds. Refer to Section~\ref{sec:pseudo-quality} for complete details.}
    \label{fig:pseudoquality}
\end{figure}

To further confirm our findings, we conduct a qualitative study in the data-scarce setup, where we compute the ROUGE scores of the final 5 pseudo-labels for each method against the respective ground truth summaries, and Figure~\ref{fig:pseudoquality} shows the mean ROUGE-2 of the five selected pseudo-labels.
To clarify, we obtained the ``Oracle" results by directly selecting the final 5 pseudo-labels using ROUGE-2 scores computed against the ground truth.
We note a stark difference between ``Confidence" and ``Oracle," which shows that relying solely on teacher confidence consistently leads to a selection of low-quality pseudo-labels. 
Combining GPT-4 score with teacher confidence is effective (``Confidence + G-4 score"), and adding the GPT-4 relabeling greatly boosts the quality of selected pseudo-labels (``Ours").

\subsection{Effect of Relabeling}
Referring to Tables~\ref{tab:full_results} and ~\ref{tab:sup_vs_semisup}, we observe that relabeling with LLMs leads to a significant boost in the summarization performance in terms of both ROUGE scores and L-Eval.
When using BERT$_{base}$ as the backbone, we note that the ROUGE-1 improves from 52.3 to 58.9 on the TweetSumm dataset, 25.2 to 26.1 on the WikiHow dataset, and 27.7 to 29.1 on the ArXiv/PubMed dataset.
GPT-4 relabeling is also effective when using teacher confidence without GPT-4 score (``Confidence" v/s ``Confidence + G-4 relabeling").
Our previous qualitative study also supports these results, showing that relabeling improves the quality of pseudo-labels.
We observe similar trends when using DistilBERT$_{base}$ as PreSumm's backbone and LLaMA-3 instead of GPT-4.
When using 500 labels, we note boosts in performance but the relative scale is smaller compared to when using 50 labels.

We conduct additional testing to analyze the performance of our best- and second-best-performing models, both of which involve relabeling.
We find that the p-value < 0.016 for Welch’s test for an R-1 of 58.9 (1.4) for ``Confidence + G-4 score + G-4 relabelling" v/s 57.8 (1.2) for ``Confidence + G-4 relabeling" on the TweetSumm dataset, denoting the differences are significant.

\section{Prompt Designs} \label{prompt-design}
\renewcommand\lstlistingname{Prompt}

\begin{figure*}[t]
  \centering
  \begin{subfigure}[b]{0.4\linewidth}
    \centering
    \includegraphics[width=\linewidth]{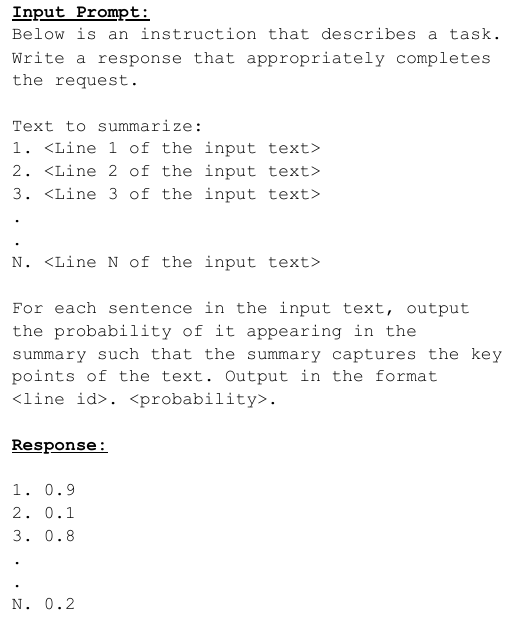}  
    \caption{\textbf{Generating pseudo-labels.} We attach a line ID to each sentence in the input document and instruct the LLM to use those line IDs in its response.}
    \label{fig:gptrelabel}
  \end{subfigure}
  \hfill
  \begin{subfigure}[b]{0.53\linewidth}
    \centering
    \includegraphics[width=\linewidth]{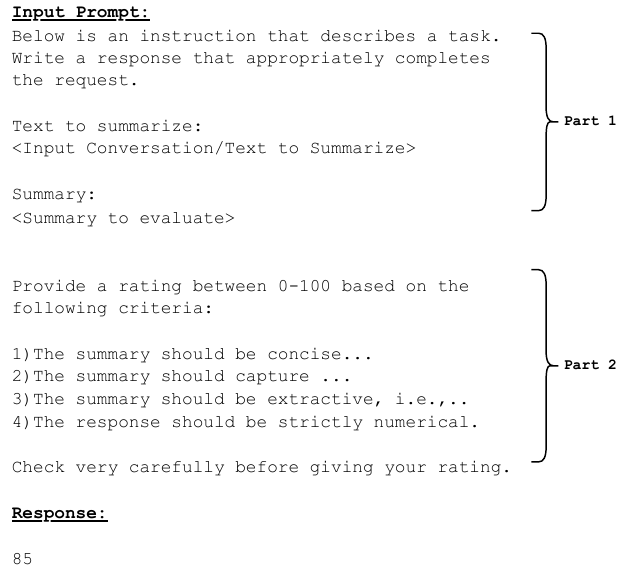}  
    \caption{\textbf{Scoring pseudo-labels.} The two-part prompt contains a text and summary pair (Part 1), and a list the evaluation criteria (Part 2). \textbf{Note:} Refer to Section~\ref{sec:score} for complete details on the evaluation criteria.}
    \label{fig:gpt4score}
  \end{subfigure}
  \caption{Different prompts used in the experiments.}
  \label{fig:whole}
\end{figure*}

In this section, we show our prompts to synthesize new documents and their summaries. 

\begin{minipage}{\textwidth}
\begin{lstlisting}[breaklines=true,label={prompt:generate},caption={Prompt used for Generating New Articles}]
### Instruction:
You are an expert data generator tasked with synthesizing new documents for a
summarization task. {dataset_description}

Below are some example documents and their summaries from a group in the dataset
(group 1):
{gp1_documents}

Below are some example documents and their summaries from another group in the dataset (group 2):
{gp2_documents}

Given the above documents, follow these instructions:
* Synthesize a new document that follows a similar format to the examples provided.
* The document should contain {document_size}.
* The document should be coherent and relevant to the topic.
* The document should be original and not copied from the examples.
* Ensure that the document covers {alpha}% topics from group 1 and 1 - {alpha}% topics from group 2.
* Wrap your response in the <document></document> tags.
* Do NOT include anything else like the examples in your output.

### Response:
\end{lstlisting}
\begin{lstlisting}[breaklines=true,label={prompt:score},caption={Prompt used for Scoring a Generated Summary}]
Below is an instruction that describes a task. Write a response that appropriately
completes the request.

### Instruction:
Given the document:
{document}

Provided Summary:
{summary}

Follow these instructions when writing your response:
* On a scale of 1-10, provide a numerical rating for the provided summary, with 10 denoting that the provided answer perfectly surmises the main points of the document.
* Your response should contain only the numerical rating. DO NOT include anything else like the provided answer, the ground truth answer, or an explanation of your rating scale in your response.
* Wrap your numerical rating inside <rating></rating> tags.
* Check very carefully before answering.
* Follow the output format as shown in the example below:
Example response:
<rating>7</rating>

### Response:
\end{lstlisting}
\end{minipage}

\begin{minipage}{\textwidth}
\begin{lstlisting}[breaklines=true,label={prompt:summarize},caption={Prompt used for Summarizing an Article in {\method}}]
Below is an instruction that describes a task. Write a response that appropriately
completes the request.

### Instruction:
You are an expert data annotator tasked with summarizing documents for a summarization task. {dataset_description}

Below are some example documents and their summaries from a group in the dataset (group 1):
{gp1_documents}

Below are some example documents and their summaries from another group in the dataset (group 2):
{gp2_documents}

A document is composed of the following sentences:
{sentences}

Given the sentences above:
* You are to construct an extractive summary for the document by selecting some sentences from above.
* The summary captures the main points of the article.
* Now, output the probability of a sentence being included in the summary.
* Do NOT include anything else like the sentence in your output.
* Output your probabilities in the format <line id>. <probability>. Refer to the example below:
1. 0.73
2. 0.65
3. 0.95
etc.

### Response:
\end{lstlisting}
\end{minipage}

\end{document}